\pdfoutput=1
\PassOptionsToPackage{unicode}{hyperref}
\PassOptionsToPackage{hyphens}{url}
\PassOptionsToPackage{dvipsnames,svgnames,x11names}{xcolor}
\documentclass[
  11pt,
  a4paper,
]{article}
\usepackage{xcolor}
\usepackage[a4paper,margin=1.15in]{geometry}
\usepackage{amsmath,amssymb}
\usepackage{iftex}
\ifPDFTeX
  \usepackage[T1]{fontenc}
  \usepackage[utf8]{inputenc}
  \usepackage{textcomp} 
\else 
  \usepackage{unicode-math} 
  \defaultfontfeatures{Scale=MatchLowercase}
  \defaultfontfeatures[\rmfamily]{Ligatures=TeX,Scale=1}
\fi
\usepackage{lmodern}
\ifPDFTeX\else
\fi
\IfFileExists{upquote.sty}{\usepackage{upquote}}{}
\IfFileExists{microtype.sty}{
  \usepackage[]{microtype}
  \UseMicrotypeSet[protrusion]{basicmath} 
}{}
\makeatletter
\@ifundefined{KOMAClassName}{
  \IfFileExists{parskip.sty}{%
    \usepackage{parskip}
  }{
    \setlength{\parindent}{0pt}
    \setlength{\parskip}{6pt plus 2pt minus 1pt}}
}{
  \KOMAoptions{parskip=half}}
\makeatother
\usepackage{color}
\usepackage{fancyvrb}

\DefineVerbatimEnvironment{Highlighting}{Verbatim}{commandchars=\\\{\}}
\newenvironment{Shaded}{}{}

\newcommand{\AttributeTok}[1]{\textcolor[rgb]{0.49,0.56,0.16}{#1}}

\newcommand{\ExtensionTok}[1]{#1}

\newcommand{\NormalTok}[1]{#1}

\usepackage{longtable,booktabs,array}
\usepackage{calc} 
\usepackage{etoolbox}
\makeatletter
\patchcmd\longtable{\par}{\if@noskipsec\mbox{}\fi\par}{}{}
\makeatother
\IfFileExists{footnotehyper.sty}{\usepackage{footnotehyper}}{\usepackage{footnote}}
\makesavenoteenv{longtable}
\usepackage{graphicx}
\makeatletter
\newsavebox\pandoc@box
\newcommand*\pandocbounded[1]{
  \sbox\pandoc@box{#1}%
  \Gscale@div\@tempa{\textheight}{\dimexpr\ht\pandoc@box+\dp\pandoc@box\relax}%
  \Gscale@div\@tempb{\linewidth}{\wd\pandoc@box}%
  \ifdim\@tempb\p@<\@tempa\p@\let\@tempa\@tempb\fi
  \ifdim\@tempa\p@<\p@\scalebox{\@tempa}{\usebox\pandoc@box}%
  \else\usebox{\pandoc@box}%
  \fi%
}
\def\fps@figure{htbp}
\makeatother
\providecommand{\tightlist}{%
  \setlength{\itemsep}{0pt}\setlength{\parskip}{0pt}}
\usepackage{setspace}
\usepackage{titlesec}
\usepackage{caption}
\usepackage{enumitem}
\usepackage{fancyhdr}
\usepackage{float}

\titleformat{\section}
  {\normalfont\Large\bfseries\raggedright}
  {\thesection.}{0.65em}{}
\titleformat{\subsection}
  {\normalfont\large\bfseries\raggedright}
  {\thesubsection}{0.6em}{}
\titleformat{\subsubsection}
  {\normalfont\normalsize\bfseries\raggedright}
  {\thesubsubsection}{0.55em}{}
\titlespacing*{\section}{0pt}{1.7ex plus .35ex}{0.85ex plus .1ex}
\titlespacing*{\subsection}{0pt}{1.35ex plus .25ex}{0.55ex plus .08ex}
\titlespacing*{\subsubsection}{0pt}{1.1ex plus .2ex}{0.4ex plus .05ex}

\setlist[itemize]{leftmargin=1.35em,itemsep=0.22em,topsep=0.35em,parsep=0.08em}
\setlist[enumerate]{leftmargin=1.55em,itemsep=0.32em,topsep=0.4em,parsep=0.1em}

\fancypagestyle{plain}{%
  \fancyhf{}%
  \fancyhead[L]{\small\itshape Fully Interpretable Minimal Transformers}%
  \fancyhead[R]{\small\thepage}%
}

\usepackage{tabularx}
\usepackage{xltabular}
\usepackage{seqsplit}

\date{}

\renewcommand{\tightlist}{}

\IfFileExists{fvextra.sty}{%
  \usepackage{fvextra}
  \DefineVerbatimEnvironment{Highlighting}{Verbatim}{%
    commandchars=\\\{\},
    breaklines=true,
    fontsize=\small,
    frame=single,
    framesep=5pt,
    rulecolor=\color{black!18}%
  }
}{}
\makeatletter
\def\verbatim@font{\normalfont\ttfamily\small}
\makeatother

\makeatletter
\@ifundefined{pandoc@box}{}{%
  \renewcommand*\pandocbounded[1]{%
    \sbox\pandoc@box{#1}%
    \Gscale@div\@tempa{0.70\textheight}{\dimexpr\ht\pandoc@box+\dp\pandoc@box\relax}%
    \Gscale@div\@tempb{\linewidth}{\wd\pandoc@box}%
    \ifdim\@tempb\p@<\@tempa\p@\let\@tempa\@tempb\fi
    \ifdim\@tempa\p@<\p@\scalebox{\@tempa}{\usebox\pandoc@box}%
    \else\usebox{\pandoc@box}%
    \fi
  }%
}
\makeatother
\makeatletter
\def\fps@figure{htbp}
\makeatother

\usepackage{bookmark}
\IfFileExists{xurl.sty}{\usepackage{xurl}}{} 
\hypersetup{
  pdftitle={Fully Interpretable Minimal Transformers: From Geometry to Algorithm},
  pdfauthor={Raneem Mahajne* and Toviah Moldwin*},
  colorlinks=true,
  linkcolor={black},
  filecolor={Maroon},
  citecolor={Blue},
  urlcolor={blue},
  pdfcreator={LaTeX via pandoc}}

\title{Fully Interpretable Minimal Transformers: From Geometry to
Algorithm}
\author{Raneem Mahajne* and Toviah
Moldwin*\\[0.45em]{\normalsize\itshape Edmond and Lily Safra Center for
Brain Sciences, The Hebrew University of Jerusalem}}
\date{}

\begin{document}
\maketitle
\begin{abstract}
We present a framework for building and interpreting minimal transformer
models. By constraining a transformer's embedding dimension and head
size to \(2\), we enable full two-dimensional visualization of its
internal representations. Embeddings, query/key/value transforms,
attention outputs, residual streams, and decision boundaries can all be
seen directly. Our central claim is that the learned geometry implies an
algorithm. The arrangement of points and boundaries in \(\mathbb{R}^2\)
can be read as a step-by-step procedure. We train a transformer on a
simple task where it must produce the most recently observed even number
whenever the `+' operator appears in a sequence of digits. Once trained,
we visually walk through every step of the transformer's computation. We
show how the model embeds the tokens and their respective positions in
the sequence, transforms them via the Q, K, and V matrices, uses the dot
product between the Q and K representations to form the attention
matrix, and uses the attention matrix to select values that move the
representation of each input token embedding to the region of the domain
of the output layer that will correctly predict the next token. We
introduce a suite of interpretability visualizations that make the
algorithmic interpretation of this procedure explicit. Our framework
offers a pedagogical and experimental testbed to explore how
transformers use information geometry to implement next-token
prediction.
\end{abstract}

*These authors contributed equally to this work.

\section{Introduction}\label{introduction}

Understanding how transformers (Vaswani et al., 2017) process sequences
remains a central challenge in mechanistic interpretability. Large-scale
models achieve strong performance but their internal representations are
high-dimensional and opaque. One can probe attention or activations, but
a complete picture of information flow from input to output is often
difficult to obtain.

We address this problem with minimal transformers, i.e.~models that
retain the full structure of a decoder-only transformer, but are
constrained to two-dimensional embeddings and head dimension. Every
internal state --- embeddings, queries, keys, values, attention outputs,
residual sums, and pre-softmax logit vectors --- is represented in
\(\mathbb{R}^2\). Dimensionality reduction techniques such as PCA, t-SNE
(van der Maaten \& Hinton, 2008), or UMAP (McInnes et al., 2018) are
thus not required to visualize the internal state of the model, as the
information geometry learned by the model is directly visible in the 2D
plane. We can take advantage of this direct visibility to demonstrate
how the information geometry of the transformer can be straightforwardly
interpreted as an algorithmic procedure in an illustrative task.

\subsection{Related work}\label{related-work}

Much of the work in the mechanistic interpretability literature focuses
on large language models (LLMs) and tries to interpret attention in
terms of linguistic properties (Clark et al., 2019; Vig, 2019; Wang et
al., 2022). However, such work rarely traces the full forward pass
end-to-end in a directly visualizable space; our contribution is a
complete geometric walkthrough of a transformer model. Other work takes
a more mathematical approach to deconstructing the operations performed
by each part of the transformer (Elhage et al., 2021), or indirectly
explores how information is represented internally in the transformer
model via perturbation experiments or mathematical and model engineering
techniques (Elhage et al., 2022; Bricken et al., 2023; Park et al.,
2024; Li et al., 2023; Nanda, Lee, \& Wattenberg, 2023b; Dar et al.,
2023). Recent work analyzes vision transformers through the SVD of the
query-key matrix, revealing how attention behaves differently across
layers and image regions (Pan et al., 2024).

Many interpretability studies have focused on analyzing how geometric
structure in embedding space emerges during training on arithmetic tasks
(Nanda et al., 2023a; Musat, 2024; Gromov, 2023; Zhong et al., 2023;
Welch Labs, 2025; Power et al., 2022; Quirke \& Barez, 2024; Liu et al.,
2022; Hanna et al., 2023; Stolfo et al., 2023). A different approach to
interpretability attempts to formalize transformer computations into
human-readable languages such as RASP (Weiss et al., 2021; Friedman et
al., 2023; Zhou et al., 2024; Lindner et al., 2023).

\section{Methods}\label{methods}

\subsection{Task Definition}\label{task-definition}

We adopt the ``plus-last-even'' rule, described below, as our primary
task. We chose this task because it is simple enough to be learned by a
small transformer model while still illustrating the core attention
mechanism of history-based information retrieval.

\subsubsection{The Plus-Last-Even Task}\label{the-plus-last-even-task}

The task is defined over a vocabulary \(\mathcal{V}\) of 12 tokens: the
integers \(\{0, \ldots, 10\}\) and a special operator \texttt{+}.
Sequence generation obeys:

\begin{itemize}
\tightlist
\item
  \textbf{Retrieval rule:} If \texttt{+} occurs at position \(t\), the
  output at \(t+1\) must be the most recent even integer in the
  sequence. (If there is no earlier even number, then any token can be
  chosen).
\item
  \textbf{Unconstrained positions:} All positions not immediately
  following \texttt{+} are unconstrained; any token in \(\mathcal{V}\)
  may appear. These positions provide context that the model must
  process without applying the retrieval rule.
\end{itemize}

\begin{verbatim}
5  3  8  7  +  8  10  2  4  +  4  ...
            ^               ^
       last even = 8     last even = 4
\end{verbatim}

Positions not immediately following \texttt{+} are unconstrained, so any
token may appear. The rule thus constrains only a fraction of positions;
the remainder serve as context. The model must learn to (1) identify the
\texttt{+} token, (2) scan backward through the context to locate the
most recent even number, and (3) output that number with high
probability. This is a non-trivial attention task: it requires routing
information from a variable, content-dependent past position to the
present.

\subsection{Model Architecture}\label{model-architecture}

We use a single-layer, single-head causal transformer, which is a
minimal instance of the GPT-style architecture of Radford et al.~(2018),
using scaled dot-product self-attention as in Vaswani et al.~(2017). The
model processes tokens autoregressively: at each position it conditions
on the preceding tokens within a fixed context window of \(T = 8\) and
produces a distribution over the next token. The single transformer
block contains one causal self-attention head and a feedforward network
(a two-layer MLP applied independently to each position), with a
residual connection from the block input to the output of the
self-attention sub-layer and a residual connection from the
self-attention output to the output of the feedforward sub-layer,
followed by a linear language-model head that maps the final hidden
state to vocabulary logits (Figure 1). Table 1 lists all
hyperparameters.

\begin{figure}
\centering
\includegraphics[width=\linewidth,height=0.78\textheight,keepaspectratio,alt={Figure 1. Model architecture. A one-layer, single-head causal transformer with embedding dimension d = 2, context window T = 8, and a 12-token vocabulary. The token embeddings \textbackslash mathbf\{X\} and positional embeddings \textbackslash mathbf\{P\} of the T positions (each 8 \textbackslash times 2, one row per position) are summed to give \textbackslash mathbf\{E\} = \textbackslash mathbf\{X\} + \textbackslash mathbf\{P\}. Self-attention (yellow) projects \textbackslash mathbf\{E\} into queries, keys and values, \textbackslash mathbf\{Q\} = \textbackslash mathbf\{E\}\textbackslash mathbf\{W\}\_Q\^{}\textbackslash top, \textbackslash mathbf\{K\} = \textbackslash mathbf\{E\}\textbackslash mathbf\{W\}\_K\^{}\textbackslash top and \textbackslash mathbf\{V\} = \textbackslash mathbf\{E\}\textbackslash mathbf\{W\}\_V\^{}\textbackslash top, each with a 2 \textbackslash times 2 weight matrix. The scaled scores \textbackslash mathbf\{Q\}\textbackslash mathbf\{K\}\^{}\textbackslash top/\textbackslash sqrt\{d\_k\} are causally masked and softmax-normalized to give the 8 \textbackslash times 8 attention weights \textbackslash mathbf\{A\}, and the attention output is \textbackslash mathbf\{A\}\textbackslash mathbf\{V\}. A residual connection (green) adds \textbackslash mathbf\{E\} to this output, giving \textbackslash mathbf\{Z\}. A position-wise feed-forward network (red; Linear(2,32), ReLU, Linear(32,2)) is applied to \textbackslash mathbf\{Z\}, and a second residual connection (blue) adds \textbackslash mathbf\{Z\} back, giving the final hidden states \textbackslash mathbf\{H\}. The language-model head (green; Linear(2,12)) followed by a softmax maps each row of \textbackslash mathbf\{H\} to a probability distribution over the next token (8 \textbackslash times 12). The figure shows the shape of each matrix next to it.}]{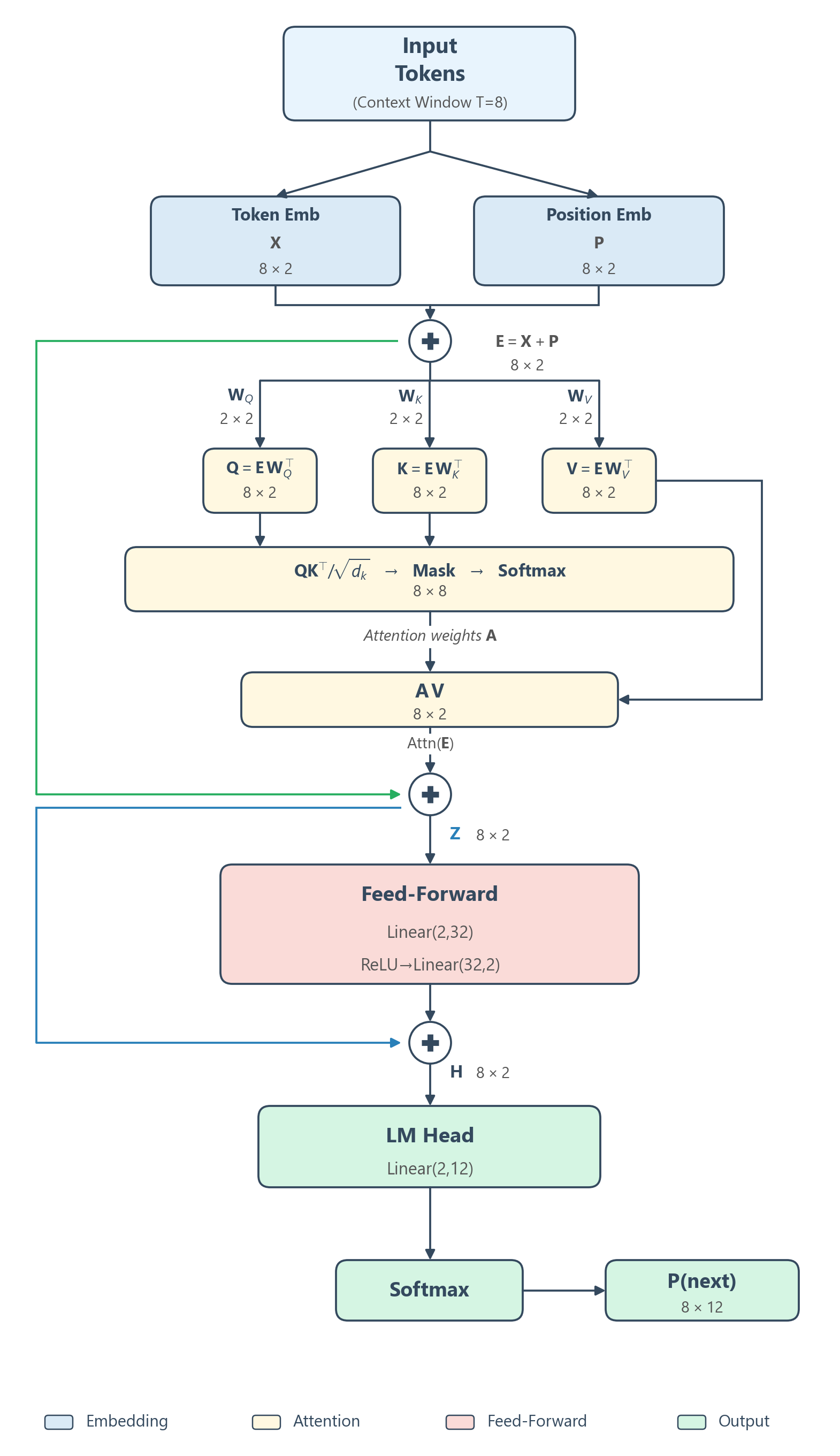}
\caption[\textbf{Figure 1.} Model architecture. A one-layer, single-head
causal transformer with embedding dimension \(d = 2\), context window
\(T = 8\), and a 12-token vocabulary. The token embeddings
\(\mathbf{X}\) and positional embeddings \(\mathbf{P}\) of the \(T\)
positions (each \(8 \times 2\), one row per position) are summed to give
\(\mathbf{E} = \mathbf{X} + \mathbf{P}\). Self-attention (yellow)
projects \(\mathbf{E}\) into queries, keys and values,
\(\mathbf{Q} = \mathbf{E}\mathbf{W}_Q^\top\),
\(\mathbf{K} = \mathbf{E}\mathbf{W}_K^\top\) and
\(\mathbf{V} = \mathbf{E}\mathbf{W}_V^\top\), each with a \(2 \times 2\)
weight matrix. The scaled scores
\(\mathbf{Q}\mathbf{K}^\top/\sqrt{d_k}\) are causally masked and
softmax-normalized to give the \(8 \times 8\) attention weights
\(\mathbf{A}\), and the attention output is \(\mathbf{A}\mathbf{V}\). A
residual connection (green) adds \(\mathbf{E}\) to this output, giving
\(\mathbf{Z}\). A position-wise feed-forward network (red; Linear(2,32),
ReLU, Linear(32,2)) is applied to \(\mathbf{Z}\), and a second residual
connection (blue) adds \(\mathbf{Z}\) back, giving the final hidden
states \(\mathbf{H}\). The language-model head (green; Linear(2,12))
followed by a softmax maps each row of \(\mathbf{H}\) to a probability
distribution over the next token (\(8 \times 12\)). The figure shows the
shape of each matrix next to it.]{\textbf{Figure 1.} Model architecture.
A one-layer, single-head causal transformer with embedding dimension
\(d = 2\), context window \(T = 8\), and a 12-token vocabulary. The
token embeddings \(\mathbf{X}\) and positional embeddings \(\mathbf{P}\)
of the \(T\) positions (each \(8 \times 2\), one row per position) are
summed to give \(\mathbf{E} = \mathbf{X} + \mathbf{P}\). Self-attention
(yellow) projects \(\mathbf{E}\) into queries, keys and values,
\(\mathbf{Q} = \mathbf{E}\mathbf{W}_Q^\top\),
\(\mathbf{K} = \mathbf{E}\mathbf{W}_K^\top\) and
\(\mathbf{V} = \mathbf{E}\mathbf{W}_V^\top\), each with a \(2 \times 2\)
weight matrix. The scaled scores
\(\mathbf{Q}\mathbf{K}^\top/\sqrt{d_k}\) are causally masked and
softmax-normalized to give the \(8 \times 8\) attention weights
\(\mathbf{A}\), and the attention output is \(\mathbf{A}\mathbf{V}\). A
residual connection (green) adds \(\mathbf{E}\) to this output, giving
\(\mathbf{Z}\). A position-wise feed-forward network (red; Linear(2,32),
ReLU, Linear(32,2)) is applied to \(\mathbf{Z}\), and a second residual
connection (blue) adds \(\mathbf{Z}\) back, giving the final hidden
states \(\mathbf{H}\). The language-model head (green; Linear(2,12))
followed by a softmax maps each row of \(\mathbf{H}\) to a probability
distribution over the next token (\(8 \times 12\)). The figure shows the
shape of each matrix next to it.}
\end{figure}

\begin{longtable}[]{@{}ll@{}}
\caption{Model hyperparameters.}\label{tbl:hyperparams}\tabularnewline
\toprule\noalign{}
Parameter & Value \\
\midrule\noalign{}
\endfirsthead
\toprule\noalign{}
Parameter & Value \\
\midrule\noalign{}
\endhead
\bottomrule\noalign{}
\endlastfoot
\(n_{\mathrm{embed}}\) & 2 \\
Block size (\(T\)) & 8 \\
Number of heads & 1 \\
Head size (\(d_k\)) & 2 \\
Vocabulary size (\(V\)) & 12 (integers 0--10, operator +) \\
Feed-forward hidden size & \(16 \times n_{\mathrm{embed}} = 32\) \\
\end{longtable}

\textbf{Token embedding:} At position \(i\) (positions are indexed from
\(0\) to \(T-1\)), the token has id \(t_i\); we denote its \textbf{token
embedding} by \(\mathbf{x}_i \in \mathbb{R}^2\). Stacking the token
embeddings of all \(T\) tokens in the context window as rows gives the
matrix \(\mathbf{X} \in \mathbb{R}^{T \times 2}\), whose row \(i\) is
\(\mathbf{x}_i^\top\).

\textbf{Positional embedding:} The embedding of position \(i\) is
\(\mathbf{p}_i \in \mathbb{R}^2\). Stacking the positional embeddings of
all \(T\) positions in the context window as rows gives the matrix
\(\mathbf{P} \in \mathbb{R}^{T \times 2}\), whose row \(i\) is
\(\mathbf{p}_i^\top\).

The \textbf{combined embedding} (input to attention) at position \(i\)
is \[
\mathbf{e}_i = \mathbf{x}_i + \mathbf{p}_i.
\] For the whole context window, the combined embeddings are stacked as
rows of the matrix \[
\mathbf{E} = \mathbf{X} + \mathbf{P} \in \mathbb{R}^{T \times n_{\mathrm{embed}}},
\] whose row \(i\) is \(\mathbf{e}_i^\top\).

\textbf{Self-attention.} Attention operates on the combined-embedding
matrix \(\mathbf{E}\). Learned projections
\(\mathbf{W}_Q, \mathbf{W}_K, \mathbf{W}_V \in \mathbb{R}^{d_k \times n_{\mathrm{embed}}}\)
map each embedding into Q, K, and V space (here
\(d_k = n_{\mathrm{embed}} = 2\)). Per position, query, key, and value
\textbf{column} vectors are
\(\mathbf{q}_i = \mathbf{W}_Q \mathbf{e}_i\),
\(\mathbf{k}_i = \mathbf{W}_K \mathbf{e}_i\), and
\(\mathbf{v}_i = \mathbf{W}_V \mathbf{e}_i\) in \(\mathbb{R}^{d_k}\).
Stacking them as \textbf{rows} gives \[
\mathbf{Q} = \mathbf{E} \mathbf{W}_Q^\top, \quad \mathbf{K} = \mathbf{E} \mathbf{W}_K^\top, \quad \mathbf{V} = \mathbf{E} \mathbf{W}_V^\top,
\] each in \(\mathbb{R}^{T \times d_k}\) (sequence length \(\times\)
head size): row \(i\) of \(\mathbf{Q}\), \(\mathbf{K}\), and
\(\mathbf{V}\) is \(\mathbf{q}_i^\top\), \(\mathbf{k}_i^\top\), and
\(\mathbf{v}_i^\top\), respectively.

The \textbf{attention matrix} \(\mathbf{A} \in \mathbb{R}^{T \times T}\)
is calculated as a row-wise causal softmax of
\(\mathbf{Q}\mathbf{K}^\top/\sqrt{d_k}\): for each \(i\),
\((A_{i0},\ldots,A_{ii}) = \mathrm{softmax}(\mathbf{q}_i^\top\mathbf{k}_0/\sqrt{d_k},\ldots,\mathbf{q}_i^\top\mathbf{k}_i/\sqrt{d_k})\).
We employ causal masking by setting \(A_{ij}=0\) for \(j>i\) to ensure
that tokens can only attend to tokens that precede them in the sequence.
We denote row \(i\) of \(\mathbf{A}\) by \(\boldsymbol{\alpha}_i\), and
its entries satisfy \(\alpha_{ij}=A_{ij}\), where \(\alpha_{ij}\) is the
attention weight from position \(i\) to position \(j\).

The \textbf{attention output} combines the value rows using the
attention weights: \[
\mathrm{Attn}(\mathbf{E}) = \mathbf{A} \mathbf{V} \in \mathbb{R}^{T \times d_k},
\] whose row \(i\) is \(\mathrm{Attn}(\mathbf{E})_i^\top\), the
attention output at position \(i\). With \(\boldsymbol{\alpha}_i\) the
\(i\)-th row of \(\mathbf{A}\), \[
\mathrm{Attn}(\mathbf{E})_i^\top = (\mathbf{A} \mathbf{V})_{i,:} = \boldsymbol{\alpha}_i \mathbf{V}.
\] Equivalently, with \(\alpha_{ij} = A_{ij}\), \[
\mathrm{Attn}(\mathbf{E})_i = \sum_{j=0}^{i} \alpha_{ij} \mathbf{v}_j.
\] Thus \(\mathrm{Attn}(\mathbf{E})_i\) is the ``message'' delivered to
position \(i\) by the attention mechanism.

\textbf{First residual.} The representation passed to the feedforward
block is the sum of the combined embedding and the attention output. We
denote this \textbf{first-residual state} by \(\mathbf{z}_i\): \[
\mathbf{z}_i = \mathbf{e}_i + \mathrm{Attn}(\mathbf{E})_i.
\] Stacking over positions, the first-residual states form the matrix \[
\mathbf{Z} = \mathbf{E} + \mathrm{Attn}(\mathbf{E}) \in \mathbb{R}^{T \times n_{\mathrm{embed}}},
\] whose row \(i\) is \(\mathbf{z}_i^\top\) (the sum is defined because
\(d_k = n_{\mathrm{embed}}\)). So \(\mathbf{z}_i\) is the quantity that
is passed into the feedforward network. (Note that the feedforward
network itself can be interpreted as a key-value operation over its
inputs; see Geva et al., 2021, 2022).

\textbf{Second residual.} The feedforward network (FFN) is applied to
\(\mathbf{z}_i\), and its output is added back to \(\mathbf{z}_i\) (the
second residual connection). The resulting \textbf{hidden state} passed
to the LM head is \[
\mathbf{h}_i = \mathbf{z}_i + \mathrm{FFN}(\mathbf{z}_i).
\] Stacking over positions, the hidden states form the matrix \[
\mathbf{H} = \mathbf{Z} + \mathrm{FFN}(\mathbf{Z}) \in \mathbb{R}^{T \times n_{\mathrm{embed}}},
\] where the FFN is applied to each row of \(\mathbf{Z}\) separately,
and whose row \(i\) is \(\mathbf{h}_i^\top\).

\textbf{LM head.} The LM head produces the next-token distribution from
the hidden state \(\mathbf{h}_i\): \[
P(t_{i+1} \mid \mathbf{h}_i) = \mathrm{softmax}\!\left(\mathbf{W}_{\mathrm{lm}} \mathbf{h}_i + \mathbf{b}\right),
\] where \(\mathbf{W}_{\mathrm{lm}} \in \mathbb{R}^{V \times 2}\) and
\(\mathbf{b} \in \mathbb{R}^V\).

The forward pass can be summarized as:

\[
\begin{aligned}
\mathbf{e}_i &= \mathbf{x}_i + \mathbf{p}_i, \\
\mathbf{z}_i &= \mathbf{e}_i + \mathrm{Attn}(\mathbf{E})_i, \\
\mathbf{h}_i &= \mathbf{z}_i + \mathrm{FFN}(\mathbf{z}_i), \\
P(t_{i+1} \mid \mathbf{h}_i) &= \mathrm{softmax}\!\left(\mathbf{W}_{\mathrm{lm}} \mathbf{h}_i + \mathbf{b}\right).
\end{aligned}
\]

\subsection{Training}\label{training}

Training data consists of 2,000 sequences of length 20--50, generated by
the plus-last-even rule. At each position, the next token is drawn as
follows. With probability 0.3 the token is \texttt{+}; with probability
0.7 it is a digit chosen uniformly from \(\{0, \ldots, 10\}\). The one
exception is the position immediately after \texttt{+}: there the next
token is fixed to the most recent even number in the prefix (the rule
target). Thus in the training distribution, \texttt{+} has marginal
probability 0.3, each digit has marginal probability
\(0.7 / 11 \approx 0.064\), and every position following \texttt{+} is a
constrained label.

The model is optimized with AdamW (\(\beta_1 = 0.9\),
\(\beta_2 = 0.999\), weight decay \(10^{-2}\), PyTorch defaults) using
standard next-token cross-entropy loss. Table 2 lists all training
hyperparameters. Checkpoints are saved every 100 steps (200 total),
enabling the construction of training-evolution animations. In our
setup, a full run of 20,000 steps took on the order of 10--15 minutes on
a laptop CPU (no GPU).

\begin{longtable}[]{@{}ll@{}}
\caption{Training hyperparameters.}\label{tbl:training}\tabularnewline
\toprule\noalign{}
Parameter & Value \\
\midrule\noalign{}
\endfirsthead
\toprule\noalign{}
Parameter & Value \\
\midrule\noalign{}
\endhead
\bottomrule\noalign{}
\endlastfoot
Optimizer & AdamW \\
Learning rate & \(10^{-3}\) \\
Batch size & 8 \\
Training steps & 20,000 \\
Evaluation interval & every 200 steps \\
Evaluation iterations & 50 batches \\
Checkpoint interval & every 100 steps \\
\end{longtable}

\section{Results}\label{results}

\subsection{The Model Learns the Rule}\label{the-model-learns-the-rule}

We first verify that training succeeds by examining the training data,
the model's own generations, and the learning curve. Figure 2 displays
four sample training sequences as heatmaps. At each position we show
whether the next-token label is correct (green), wrong (red), or
unconstrained (gray). A position is \emph{constrained} if it immediately
follows \texttt{+} (the rule then requires the next token to be the most
recent even number); \emph{unconstrained} positions may have any next
token. All constrained positions in the training data are green,
verifying the data generator.

\begin{figure}
\centering
\pandocbounded{\includegraphics[keepaspectratio,alt={Figure 2. Sample training sequences as heatmaps. Green: constrained position (immediately after +) with the correct next token (the most recent even number). Gray: unconstrained position (any token is valid).}]{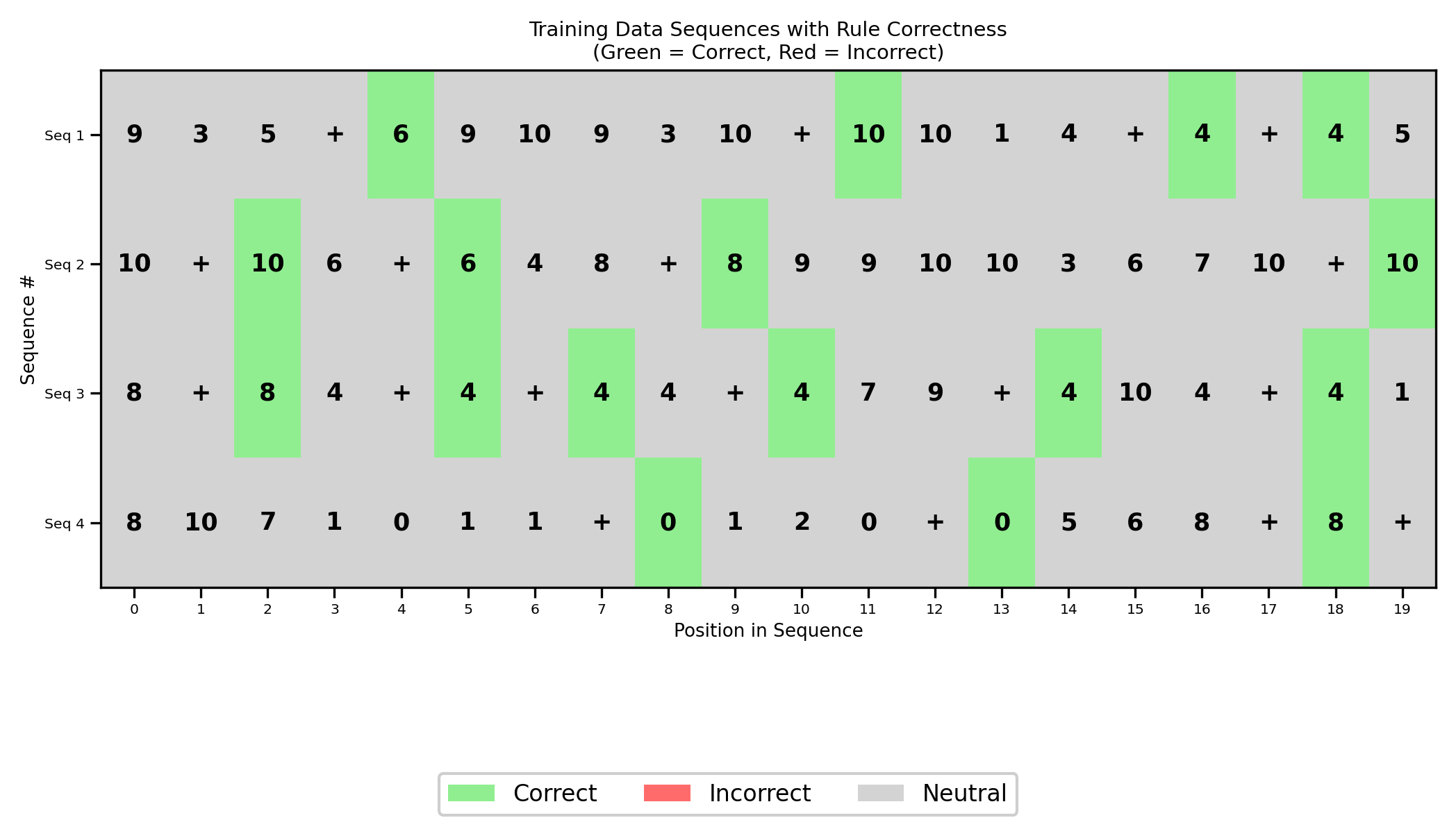}}
\caption[\textbf{Figure 2.} Sample training sequences as heatmaps.
Green: constrained position (immediately after \protect\texttt{+}) with
the correct next token (the most recent even number). Gray:
unconstrained position (any token is valid).]{\textbf{Figure 2.} Sample
training sequences as heatmaps. Green: constrained position (immediately
after \protect\texttt{+}) with the correct next token (the most recent
even number). Gray: unconstrained position (any token is valid).}
\end{figure}

Figure 3 shows the training dynamics over 20,000 steps. Cross-entropy
loss drops steeply in the first \textasciitilde2,000 steps, then
continues to decrease more gradually. \emph{Rule error} is the fraction
of constrained positions (those immediately following \texttt{+}) where
the model's top prediction is wrong; it drops from \textasciitilde90\%
(chance for a 12-token vocabulary) to near 0\%, indicating that the
model has learned the rule. Rule error on held-out batches is our
primary quantitative performance metric (Figure 3). Rule error stays
approximately flat for the first \textasciitilde7,500 steps even while
the cross-entropy loss decreases because in the initial phase of
learning, the model first learns the overall token probabilities and to
distinguish the \texttt{+} token before learning to correctly follow the
plus-last-even rule (see Movies 1--4).

\begin{figure}
\centering
\pandocbounded{\includegraphics[keepaspectratio,alt={Figure 3. Training dynamics over 20,000 steps. Left axis: cross-entropy loss. Right axis: rule error --- the fraction of constrained positions (tokens immediately following +) where the model's top predicted token is not the target (the most recent even number). Rule error near 90\% is chance level; near 0\% indicates the rule is learned.}]{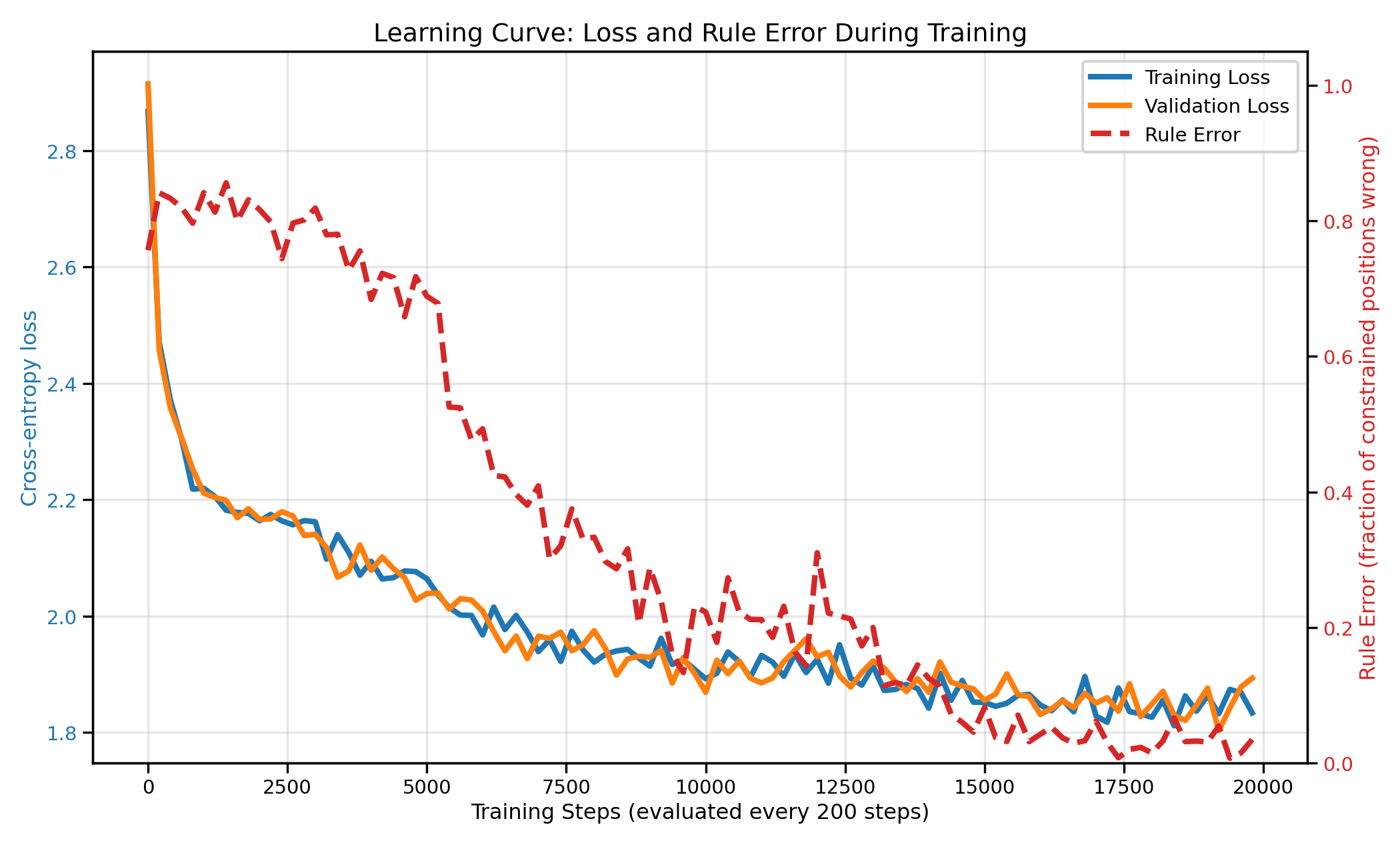}}
\caption[\textbf{Figure 3.} Training dynamics over 20,000 steps. Left
axis: cross-entropy loss. Right axis: rule error --- the fraction of
constrained positions (tokens immediately following \protect\texttt{+})
where the model's top predicted token is not the target (the most recent
even number). Rule error near 90\% is chance level; near 0\% indicates
the rule is learned.]{\textbf{Figure 3.} Training dynamics over 20,000
steps. Left axis: cross-entropy loss. Right axis: rule error --- the
fraction of constrained positions (tokens immediately following
\protect\texttt{+}) where the model's top predicted token is not the
target (the most recent even number). Rule error near 90\% is chance
level; near 0\% indicates the rule is learned.}
\end{figure}

Figure 4 compares sequences generated by the model before and after
training. The same color convention applies: at initialization (step 0),
predictions at constrained positions are essentially random --- most
cells are red. After training, nearly all constrained positions are
green: the model reliably outputs the most recent even number after
every \texttt{+}. We now proceed to show how the model implements the
``plus-last-even'' rule.

\begin{figure}
\centering
\pandocbounded{\includegraphics[keepaspectratio,alt={Figure 4. Model-generated sequences at initialization (top) vs.~after training (bottom). Green = correct at constrained positions (after +), red = wrong at constrained positions, gray = unconstrained. Top: at step 0, constrained positions are mostly red (random predictions). Bottom: after training, constrained positions are mostly green (correct ``last even'' outputs).}]{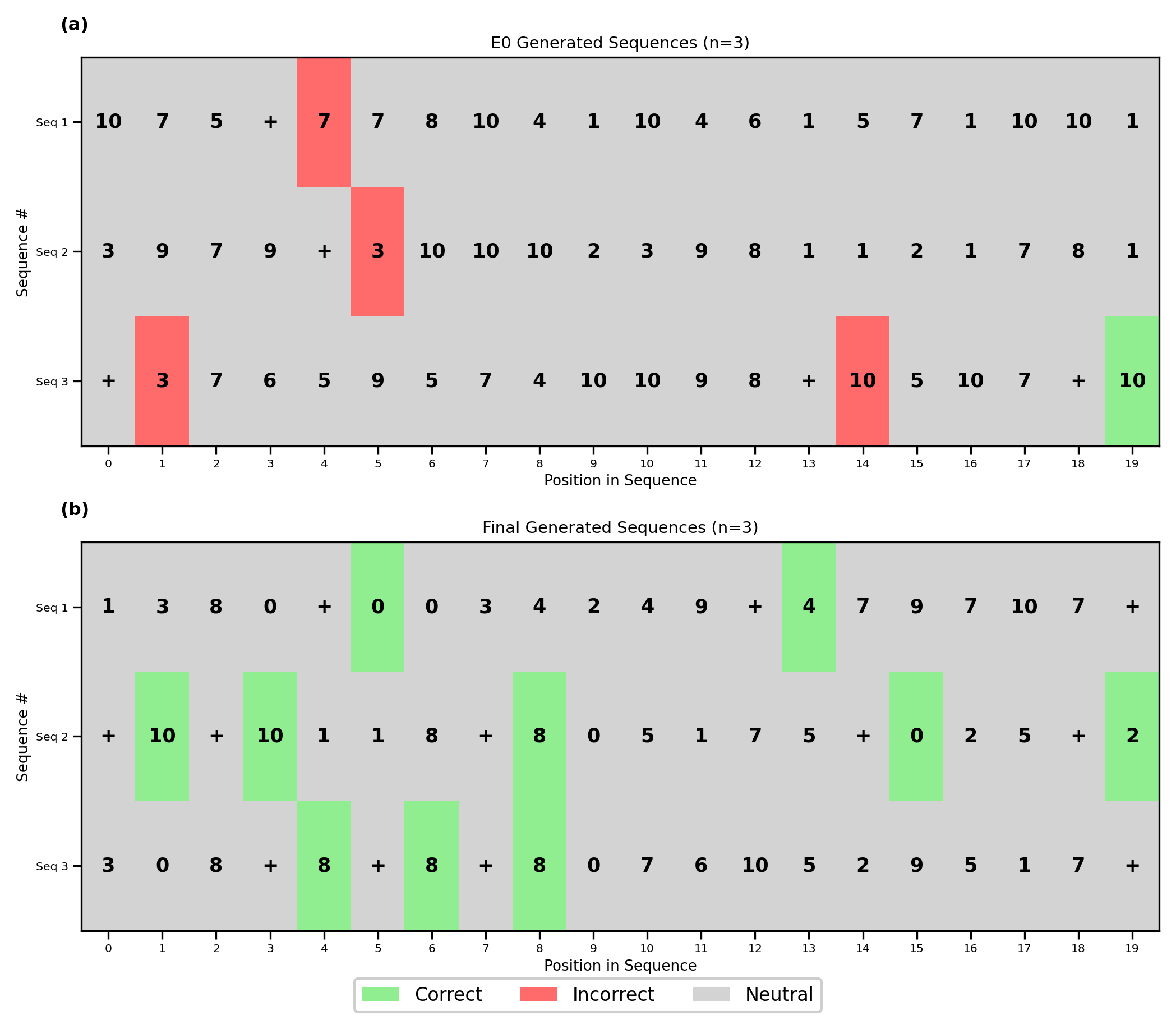}}
\caption[\textbf{Figure 4.} Model-generated sequences at initialization
(top) vs.~after training (bottom). Green = correct at constrained
positions (after \protect\texttt{+}), red = wrong at constrained
positions, gray = unconstrained. Top: at step 0, constrained positions
are mostly red (random predictions). Bottom: after training, constrained
positions are mostly green (correct ``last even''
outputs).]{\textbf{Figure 4.} Model-generated sequences at
initialization (top) vs.~after training (bottom). Green = correct at
constrained positions (after \protect\texttt{+}), red = wrong at
constrained positions, gray = unconstrained. Top: at step 0, constrained
positions are mostly red (random predictions). Bottom: after training,
constrained positions are mostly green (correct ``last even'' outputs).}
\end{figure}

\subsection{The Embedding Space: How the Model Encodes Its Vocabulary}\label{the-embedding-space-how-the-model-encodes-its-vocabulary}

The first stage of the transformer maps each input token and position to
a 2D vector. Figure 5 reveals that the learned embedding layer has
already done significant organizational work before attention is
applied. In the token embedding scatter plot (Figure 5c), the six even
numbers (0, 2, 4, 6, 8, 10) sit in a spaced formation at the top of the
plane, the five odd numbers (1, 3, 5, 7, 9) bunch together in the center
of the plane, and the \texttt{+} operator sits far from both groups as
an isolated outlier. The model has discovered that the categories
relevant to the rule --- even numbers, odd numbers, and the operator ---
should occupy geometrically distinct regions. Moreover, the `personal
space' given to the even-digit tokens, as opposed to the bunched
formation of the odd tokens, indicates that the distinct identity of the
even tokens is more important for solving this task.

The position embeddings (Figure 5d) form a ladder structure, with
\(\mathbf{p}_0\) at the bottom, \(\mathbf{p}_7\) at the top, and the
other positions arranged in ascending order in between. This orderly
arrangement allows the model to encode how far back a token is, which is
essential for identifying the most recent even number. When token and
position embeddings are summed (Figure 5e), each token fans out into
eight copies --- one per position --- shifted vertically by the position
embedding. Because the range of values of the position embeddings is
smaller than the range of the token embeddings, the `macro'-level
geometry of the summed token+position embeddings retains the
odd/even/\texttt{+} organization of the token embeddings, while the
`micro'-level geometry preserves the ladder structure of the position
embeddings.

\begin{figure}
\centering
\includegraphics[width=\linewidth,height=0.6\textheight,keepaspectratio,alt={Figure 5. Learned embeddings. (a) Token embeddings \textbackslash mathbf\{x\}\_i shown as heatmap. (b) Position embeddings \textbackslash mathbf\{p\}\_i shown as heatmap. (c) Token embeddings shown as scatterplot. Inset: zoom-in on cluster of odd digits. (d) Positions \textbackslash mathbf\{p\}\_i shown as scatterplot. (e) Token+position embeddings \textbackslash mathbf\{e\}\_i shown as scatterplot. Insets magnify the odd-digit cluster at positions 4 (blue) and 3 (orange). Positions are indicated as subscripts next to each token, e.g., +\_5 is the + token at position 5.}]{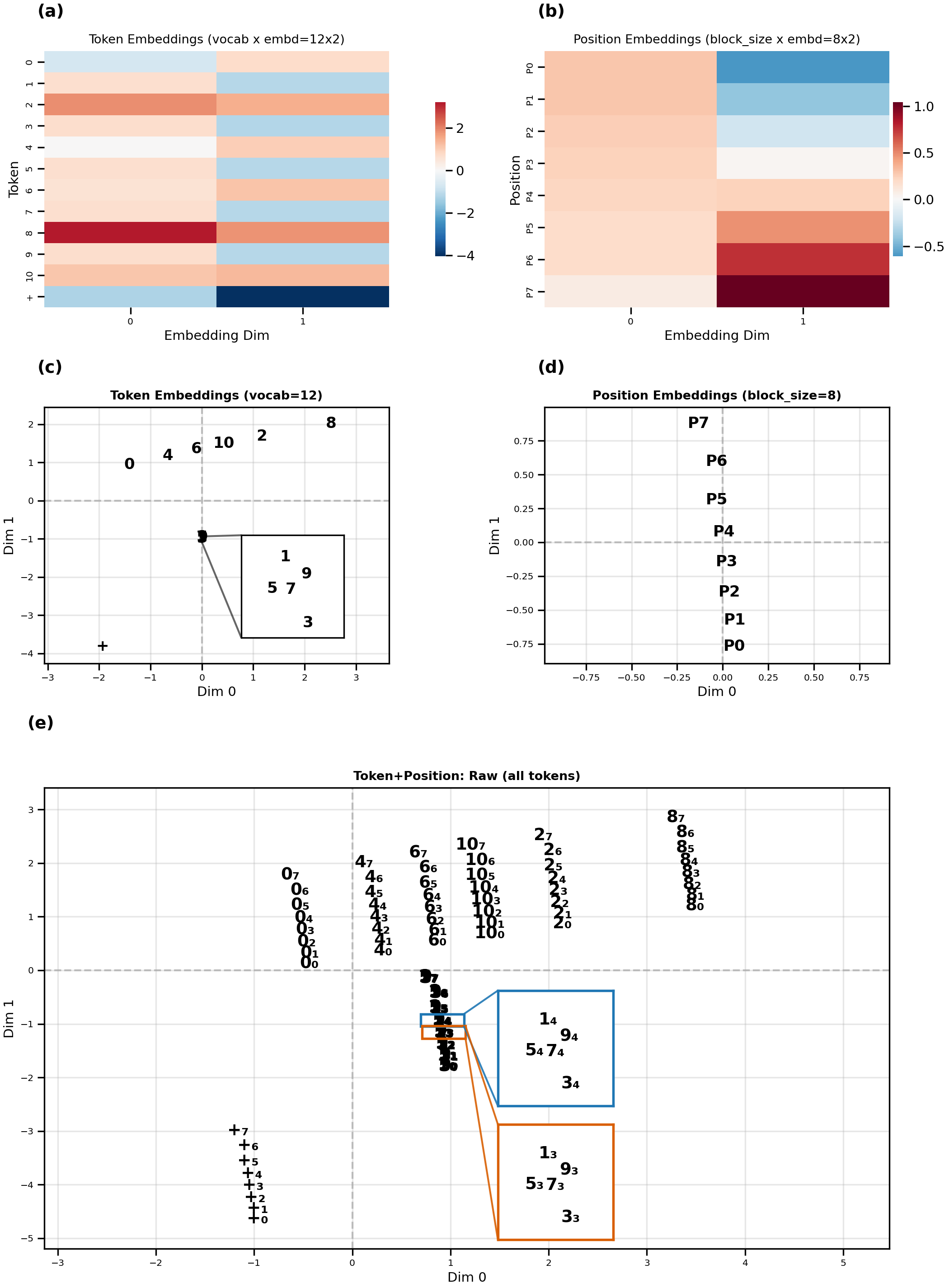}
\caption[\textbf{Figure 5.} Learned embeddings. (a) Token embeddings
\(\mathbf{x}_i\) shown as heatmap. (b) Position embeddings
\(\mathbf{p}_i\) shown as heatmap. (c) Token embeddings shown as
scatterplot. Inset: zoom-in on cluster of odd digits. (d) Positions
\(\mathbf{p}_i\) shown as scatterplot. (e) Token+position embeddings
\(\mathbf{e}_i\) shown as scatterplot. Insets magnify the odd-digit
cluster at positions 4 (blue) and 3 (orange). Positions are indicated as
subscripts next to each token, e.g., \(+_5\) is the \protect\texttt{+}
token at position 5.]{\textbf{Figure 5.} Learned embeddings. (a) Token
embeddings \(\mathbf{x}_i\) shown as heatmap. (b) Position embeddings
\(\mathbf{p}_i\) shown as heatmap. (c) Token embeddings shown as
scatterplot. Inset: zoom-in on cluster of odd digits. (d) Positions
\(\mathbf{p}_i\) shown as scatterplot. (e) Token+position embeddings
\(\mathbf{e}_i\) shown as scatterplot. Insets magnify the odd-digit
cluster at positions 4 (blue) and 3 (orange). Positions are indicated as
subscripts next to each token, e.g., \(+_5\) is the \protect\texttt{+}
token at position 5.}
\end{figure}

Movie 1 shows how the geometric structure of the embedding space evolves
over the course of learning. At step 0, all points are randomly
scattered. Within the first few thousand steps, the \texttt{+} token
rapidly migrates away from the number tokens. The even/odd split
solidifies between steps 5,000 and 10,000, and the position embedding
ladder organizes gradually throughout training.

\subsection{The Attention Mechanism: Query, Key, and Value Projections}\label{the-attention-mechanism-query-key-and-value-projections}

The combined embeddings \(\mathbf{e}_i\) alone are not sufficient for
the model to obey the plus-last-even rule, due to the rule's conditional
structure. The \texttt{+} tokens need to ``search back'' for the most
recent even number to generate the correct answer; simply knowing the
token value and its position is not enough to predict the next token.
This retrieval process is what motivates the use of the attention
mechanism (Vaswani et al., 2017).

To illustrate how the attention mechanism works, consider what happens
when the model encounters a \texttt{+} token and needs to retrieve the
most recent even number. Each combined embedding \(\mathbf{e}_i\) is
linearly transformed into three vectors: a query (\(\mathbf{q}_i\)), a
key (\(\mathbf{k}_i\)), and a value (\(\mathbf{v}_i\)). For example,
when processing \(+_7\) (the \texttt{+} token at position 7), the query
vector represents the ``search request'' asking, ``Where is the relevant
even number?'' Every other position in the sequence, including those
with even numbers, supplies its own key and value vectors. The model
computes the attention weights for the \texttt{+} token by taking the
dot product between its query (\(\mathbf{q}_i\)) and every other token's
key (\(\mathbf{k}_j\)). This produces a score indicating how strongly
the \texttt{+} should attend to each past token---ideally, giving the
highest score to the key corresponding to the most recent even number.
The value vectors (\(\mathbf{v}_j\)) determine what information can
actually be retrieved---so the value for the most recent even number
carries semantic information about that number, which is then used to
generate the correct output after \texttt{+}. In summary: the query and
key vectors let the \texttt{+} ``look back'' specifically for even
numbers (and, thanks to position embeddings, for the most recent one),
while the value vectors supply the actual information necessary to
produce the correct output.

The model applies three learned linear maps
\(\mathbf{W}_Q, \mathbf{W}_K, \mathbf{W}_V \in \mathbb{R}^{d_k \times n_{\mathrm{embed}}}\)
(here \(n_{\mathrm{embed}} = d_k = 2\)) to every combined embedding
\(\mathbf{e}_i\), producing query, key, and value column vectors in
\(\mathbb{R}^{d_k}\); for a sequence of tokens of length \(T\) these
form
\(\mathbf{Q}, \mathbf{K}, \mathbf{V} \in \mathbb{R}^{T \times d_k}\).
Figure 6 shows these transformations and their effect. Panel (a)
displays the original combined token+position embeddings (all 96 points
in embedding space). Panels (b), (c), and (d) show the same points after
projection into query space (blue), key space (red), and value space
(green).

\begin{figure}
\centering
\pandocbounded{\includegraphics[keepaspectratio,alt={Figure 6. Q, K, and V projections. (a) Original combined token+position embeddings (as in Figure 5e). (b) Query-space (blue), with \textbackslash mathbf\{W\}\_Q inset. (c) Key-space (red), with \textbackslash mathbf\{W\}\_K inset. (d) Value-space (green), with \textbackslash mathbf\{W\}\_V inset. All 96 token+position points in each panel.}]{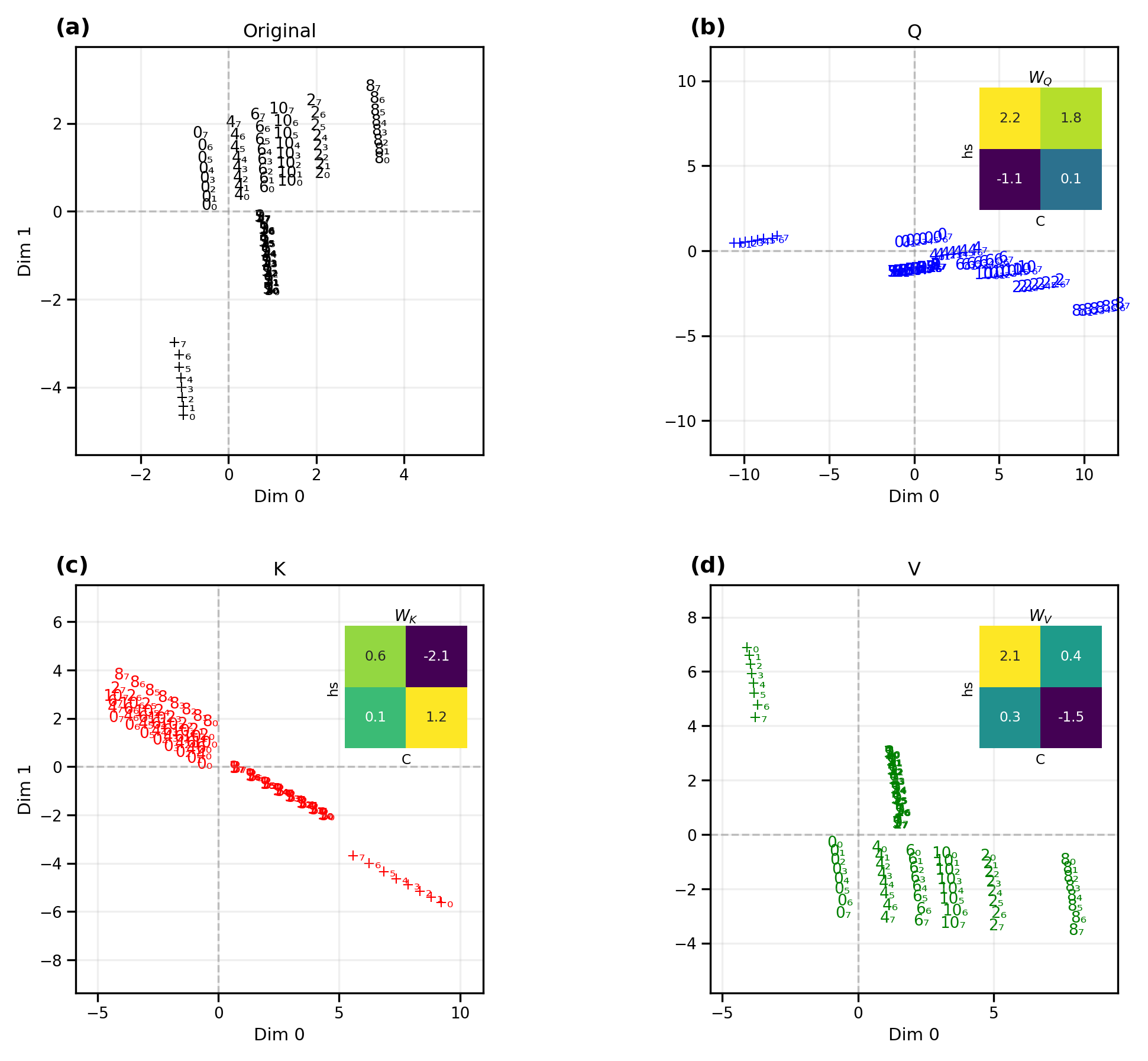}}
\caption[\textbf{Figure 6.} Q, K, and V projections. (a) Original
combined token+position embeddings (as in Figure 5e). (b) Query-space
(blue), with \(\mathbf{W}_Q\) inset. (c) Key-space (red), with
\(\mathbf{W}_K\) inset. (d) Value-space (green), with \(\mathbf{W}_V\)
inset. All 96 token+position points in each panel.]{\textbf{Figure 6.}
Q, K, and V projections. (a) Original combined token+position embeddings
(as in Figure 5e). (b) Query-space (blue), with \(\mathbf{W}_Q\) inset.
(c) Key-space (red), with \(\mathbf{W}_K\) inset. (d) Value-space
(green), with \(\mathbf{W}_V\) inset. All 96 token+position points in
each panel.}
\end{figure}

\subsection{Who Attends to Whom: The Query--Key Geometry}\label{who-attends-to-whom-the-querykey-geometry}

When the model encounters a \texttt{+}, how does it know to look back
and find the most recent even number? The answer lies in the geometry of
the query--key space. Figure 7 plots all 96 query vectors (blue) and all
96 key vectors (red) on common axes, so we can directly read off the
attention pattern from the spatial relationships.

Figure 7 (Top) makes this geometry visible. The \texttt{+} queries form
a tight single-file group well-separated from all digit queries.
Even-number keys are concentrated in the region closest to that
\texttt{+} cluster, while odd-number keys lie in a distinctly different
region, far from it. This spatial arrangement is the geometric encoding
of the rule's core requirement that the \texttt{+} token should attend
to even numbers and ignore odd numbers.

But the rule demands more than just ``attend to even numbers''. It
requires attending to the \emph{most recent} one. This is encoded in the
positional spread of each token's keys. For any given even number, keys
at later positions (closer to the query) are arranged so that they are
closer to the \texttt{+} queries than keys at earlier positions.

If we focus on a single query \(+_5\) (\texttt{+} at position 5), we can
see the dot product (green background) between that query and a key at
any point in space (Figure 7 Bottom). In principle, the dot product is
highest with even numbers at position 7; however, because of the causal
masking, queries are only allowed to look at keys that come before them
in the sequence, so we have grayed out all keys at position \(\geq 5\)
(keys at position 5 are also grayed out because if there's a \texttt{+}
in position 5, no other token could be in position 5). The \texttt{+}
query thus acts as a selective filter that picks out even-number keys
from the past context, with recency biasing the selection toward the
most recent one. Movie 4 reveals how this structure develops over the
course of training.

\begin{figure}
\centering
\pandocbounded{\includegraphics[keepaspectratio,alt={Figure 7. Query--key geometry in Q/K space. Top: joint query--key space (blue: queries; red: keys; labels show token and position). Bottom: Focus on the query +\_5 (blue); background color indicates the dot product of the +\_5 query with every point in space. Keys with position \textbackslash geq 5 are grayed due to the causal mask.}]{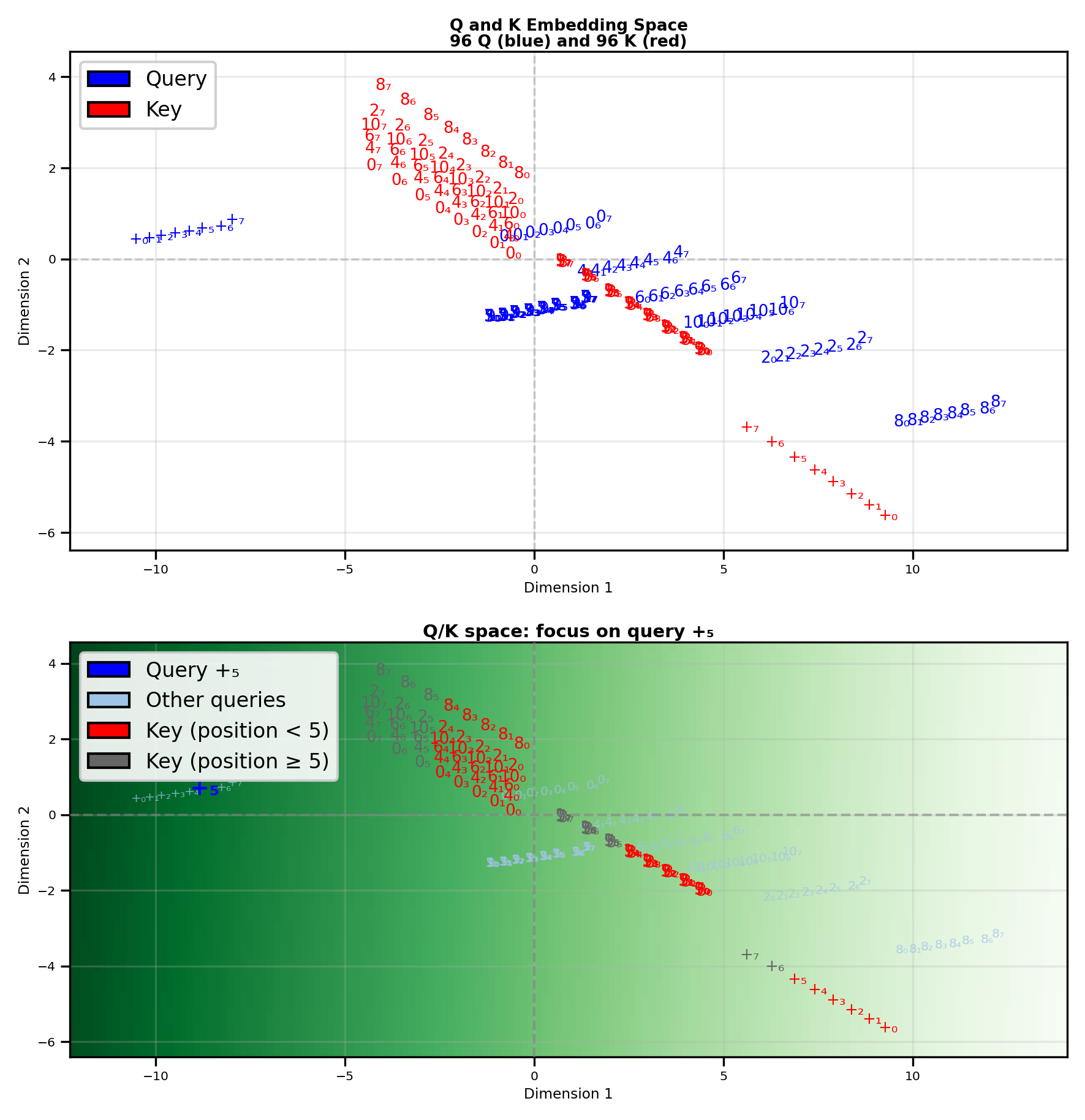}}
\caption[\textbf{Figure 7.} Query--key geometry in Q/K space.
\textbf{Top:} joint query--key space (blue: queries; red: keys; labels
show token and position). \textbf{Bottom:} Focus on the query \(+_5\)
(blue); background color indicates the dot product of the \(+_5\) query
with every point in space. Keys with position \(\geq 5\) are grayed due
to the causal mask.]{\textbf{Figure 7.} Query--key geometry in Q/K
space. \textbf{Top:} joint query--key space (blue: queries; red: keys;
labels show token and position). \textbf{Bottom:} Focus on the query
\(+_5\) (blue); background color indicates the dot product of the
\(+_5\) query with every point in space. Keys with position \(\geq 5\)
are grayed due to the causal mask.}
\end{figure}

Because Figure 8 is not tied to a specific input sequence, it is not
possible to produce a true attention matrix, which would require having
only a single token in each position. However, we can calculate the raw
query--key scores \(\mathbf{Q}\mathbf{K}^\top\), i.e.~the dot product
between every possible query vector and every possible key vector. To
focus on the rule-relevant structure, Figure 8 shows only the blocks
where the query token is \texttt{+}. Each subplot is an 8×8 grid: rows
index the position of the \texttt{+} query, and columns index the
position of the key for the indicated key token. Even-number key blocks
score consistently higher than odd-number key blocks, showing that
\texttt{+} queries score even keys more strongly across positions. (The
full all-query \(\mathbf{Q}\mathbf{K}^\top\) matrix is included in
Supplementary Figure S5). Moreover, tokens at later positions receive
higher dot product scores than tokens at earlier positions. This
confirms the observation from Figure 7 that the geometry of the
key/query space allows the \texttt{+} tokens to attend to the most
recent even tokens.

\begin{figure}
\centering
\pandocbounded{\includegraphics[keepaspectratio,alt={Figure 8. Pre-softmax query--key scores \textbackslash mathbf\{Q\}\textbackslash mathbf\{K\}\^{}\textbackslash top, restricted to + queries. Each subplot shows an 8×8 matrix of dot products: + query positions (rows) vs.~key positions (columns) for a particular key token.}]{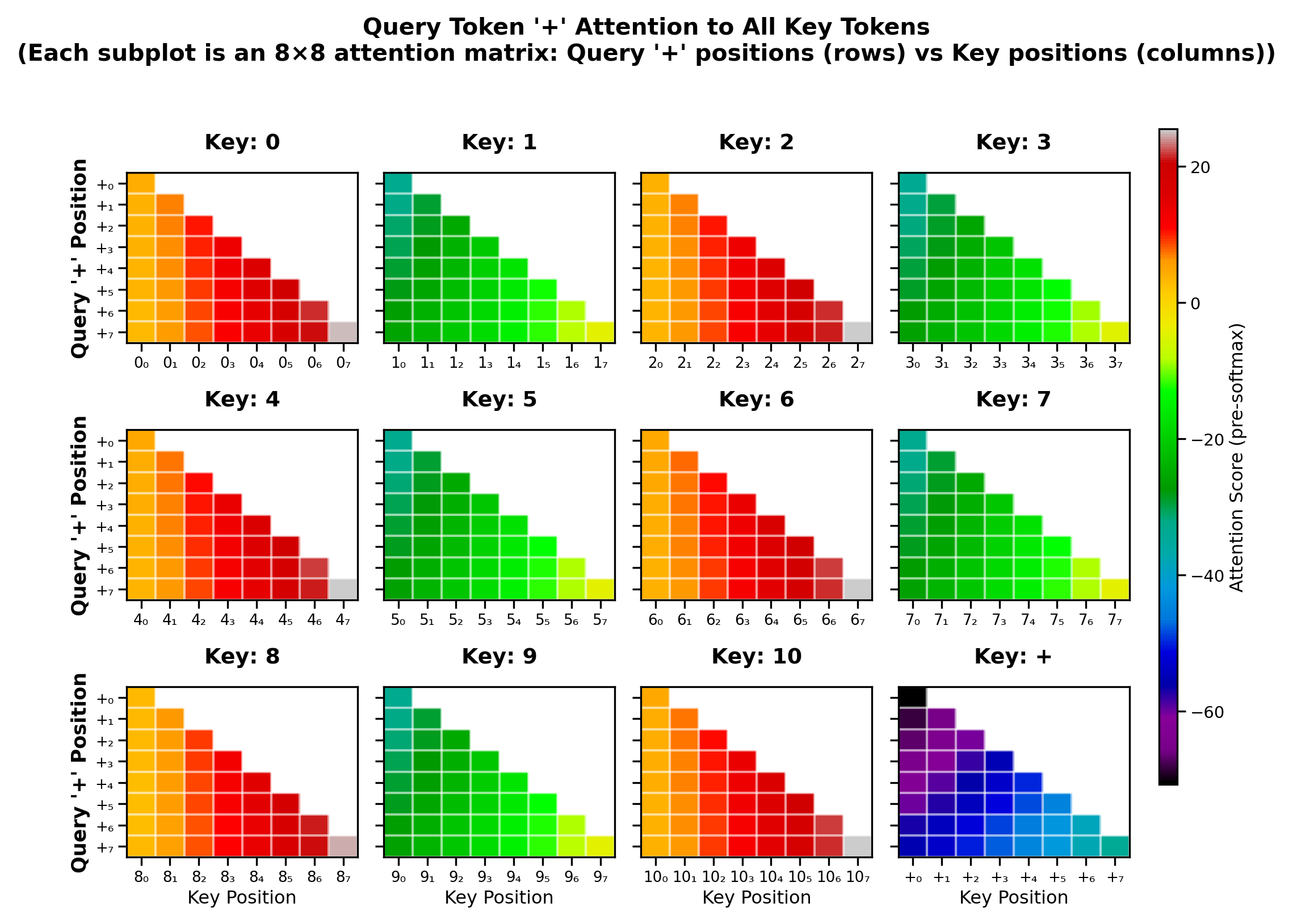}}
\caption[\textbf{Figure 8.} Pre-softmax query--key scores
\(\mathbf{Q}\mathbf{K}^\top\), restricted to \protect\texttt{+} queries.
Each subplot shows an 8×8 matrix of dot products: \protect\texttt{+}
query positions (rows) vs.~key positions (columns) for a particular key
token.]{\textbf{Figure 8.} Pre-softmax query--key scores
\(\mathbf{Q}\mathbf{K}^\top\), restricted to \protect\texttt{+} queries.
Each subplot shows an 8×8 matrix of dot products: \protect\texttt{+}
query positions (rows) vs.~key positions (columns) for a particular key
token.}
\end{figure}

\subsection{The Output Landscape: Where Representations Need to Land}\label{the-output-landscape-where-representations-need-to-land}

Before examining the value transformation, we first characterize the
output network through which all information must ultimately pass. This
will help us better appreciate the purpose of the values.

The output network (FFN, second residual, and LM head, as defined in
§2.2) maps a point \(\mathbf{z}_i \in \mathbb{R}^2\) to a probability
distribution over the vocabulary to generate the next token.
Geometrically, this means that for each possible next token (the digits
from 0 to 10 and the \texttt{+} operator), the 2D plane is partitioned
into regions with different probabilities for predicting that token.
(Because of the softmax, the probabilities across all tokens for a
particular point in the 2D plane must sum to 1.) Figure 9 makes this
partition explicit: each subplot shows the model's probability for a
specific output token across the plane.

\begin{figure}
\centering
\pandocbounded{\includegraphics[keepaspectratio,alt={Figure 9. Output probability landscape (per-unit view). Each subplot shows the softmax probability P(\textbackslash text\{next\} = \textbackslash text\{token\}) over the 2D plane for one output token (digits 0--10 and the + operator).}]{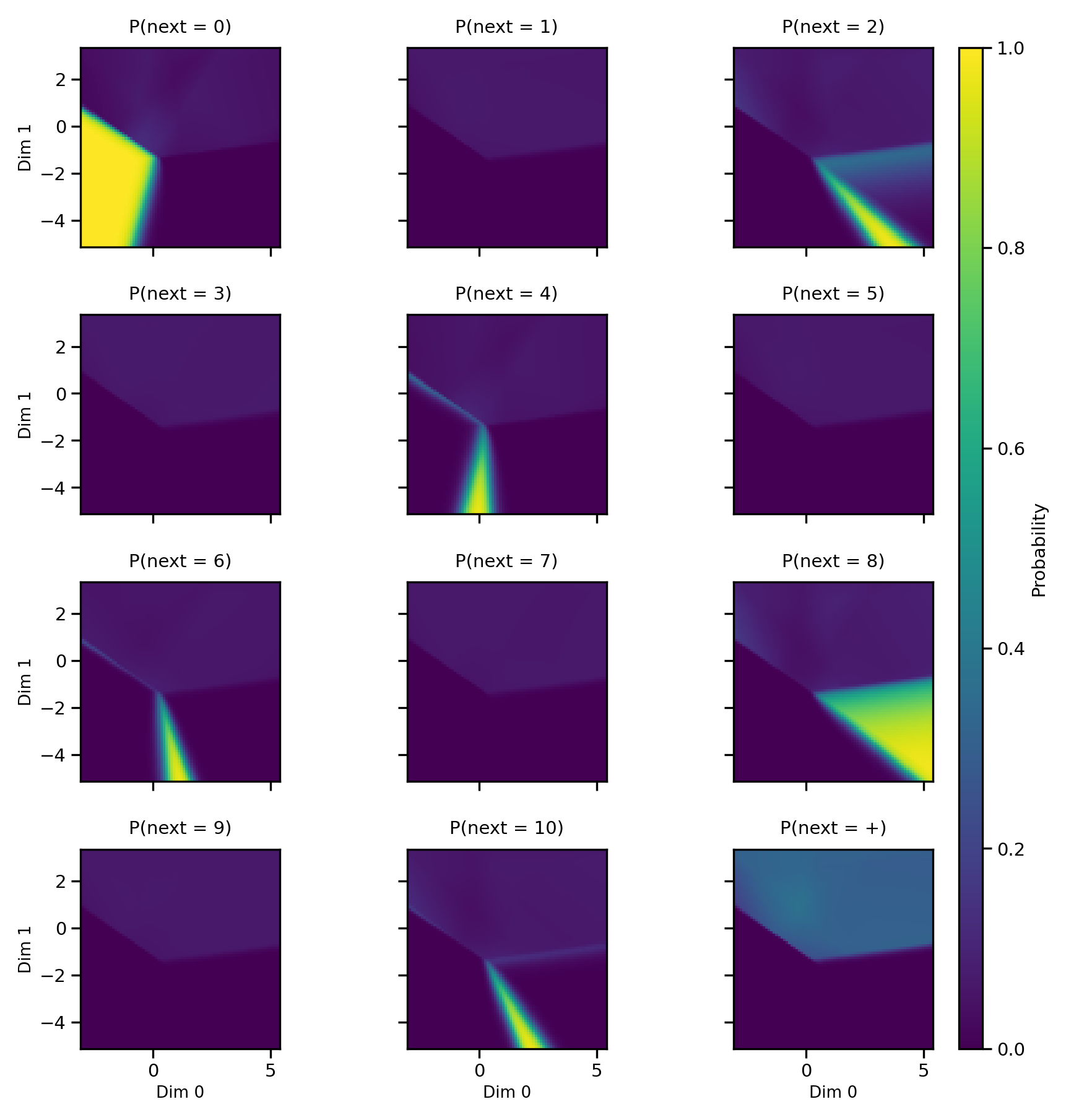}}
\caption[\textbf{Figure 9.} Output probability landscape (per-unit
view). Each subplot shows the softmax probability
\(P(\text{next} = \text{token})\) over the 2D plane for one output token
(digits 0--10 and the \protect\texttt{+} operator).]{\textbf{Figure 9.}
Output probability landscape (per-unit view). Each subplot shows the
softmax probability \(P(\text{next} = \text{token})\) over the 2D plane
for one output token (digits 0--10 and the \protect\texttt{+}
operator).}
\end{figure}

The output landscape and the embedding positions co-evolve during
training (Movie 2). At initialization, the probability landscape is
nearly uniform --- no decision boundaries exist. As the model first
learns token frequencies, broad regions form. These sharpen
progressively into the final configuration where each even number has a
well-defined, non-overlapping high-probability region.

Figure 10 provides a compact summary of the landscape shown in Figure 9.
Panel (a) shows the \emph{entropy} of the output distribution: the upper
half-plane has high entropy (many tokens are plausible), while the lower
zones have near-zero entropy (a single even number dominates). Panel (b)
shows the \emph{argmax prediction map}: at every point in the plane, the
color indicates which token the output network would predict most
strongly.

Panels (a) and (b) together reveal a clear functional decomposition of
the plane:

\begin{enumerate}
\def\labelenumi{\arabic{enumi}.}
\tightlist
\item
  \textbf{Indifference region} (upper half-plane). No single digit is
  strongly favored. The \texttt{+} operator unit shows the highest
  probability here, reflecting the training distribution: \texttt{+}
  appears with probability 0.3 and can follow any digit, so the model
  assigns it a high baseline in this unspecialized zone.
\item
  \textbf{Strong specific-prediction region} (lower half-plane). This
  area is subdivided into distinct target zones, one for each even
  integer. In each zone, one even-number unit produces a large output
  while all others are nearly zero. This region is effectively reserved
  for the final retrieval of the correct even digit after a \texttt{+}
  sign.
\end{enumerate}

Now that we understand how the output network maps a point
\(\mathbf{z}_i\) to a token prediction, we consider the individual
components of \(\mathbf{z}_i\). Recall that
\(\mathbf{z}_i = \mathbf{e}_i + \mathrm{Attn}(\mathbf{E})_i\) where
\(\mathrm{Attn}(\mathbf{E})_i = \sum_{j=0}^{i} \alpha_{ij} \mathbf{v}_j\).
In other words, \(\mathbf{z}_i\) can be thought of as being the sum of
two constituent components, the embedding \(\mathbf{e}_i\), which is
passed via the residual stream, and an attention-weighted sum of values
\(\mathbf{v}_j\). It can thus be instructive to see how each of these
components, i.e.~the embeddings \(\mathbf{e}_i\) and the values
\(\mathbf{v}_j\), are situated in the plane representing the network's
output.

Panels (c) and (d) of Figure 10 overlay the embeddings \(\mathbf{e}_i\)
and the values \(\mathbf{v}_j\), respectively, on the argmax map (panel
b) to reveal where they sit relative to the output network's decision
regions. In panel (c), the 96 combined embeddings \(\mathbf{e}_i\) (12
tokens \(\times\) 8 positions) are plotted. The digit embeddings
generally fall in the high-entropy indifference region in the upper
half-plane (visible in panel a), where no specific retrieval is
triggered. In contrast, all \texttt{+} embeddings land in the lower
half-plane --- inside the strong prediction region --- which provides a
geometric head start for the algorithm: the model has already
categorized \texttt{+} as requiring an even-number output based on token
identity and position alone. However, because the embeddings lack the
context of the rest of the string, the model can identify the ``type''
of output needed (even/odd/\texttt{+}) but cannot resolve the specific
value. This defines the attention layer's objective: it must look back
to find the most recent even number and produce a value vector that,
when added to the residual stream, nudges the representation into the
correct even-number target zone.

Panel (d) shows the value-transformed vectors
\(\mathbf{v}_j = \mathbf{W}_V \mathbf{e}_j\) for all 96 combined
token+position embeddings, overlaid on the same argmax map. We observe
that the ordering of the value vectors on the plane in Figure 10d is the
same ordering as the even-digit prediction zones in Figure 10b. This
overlay reads \(\mathbf{v}_j\) against a landscape whose native input is
\(\mathbf{z}_i\), a technique similar in spirit to the ``logit lens''
used to probe intermediate layers in large language models
(nostalgebraist, 2020; Geva et al., 2021, 2022; Belrose et al., 2023).
This illustrates that the purpose of the value vectors is to nudge the
embeddings towards a particular zone in order to produce the correct
output. However, \emph{which} value specifically will be chosen depends
on the attention matrix which requires a specific sequence. (Per-unit
heatmaps with embedding and value overlays are included in Supplementary
Figures S6 and S7.)

\begin{figure}
\centering
\includegraphics[width=\linewidth,height=0.58\textheight,keepaspectratio,alt={Figure 10. Output landscape summary (summary of Figure 9). (a) Entropy of the output distribution (nats); high in the indifference region (top half), near zero in even-number target zones (bottom half). (b) Argmax prediction map: color and annotation indicate which token has the highest predicted probability at each point. (c) Argmax map with combined embeddings \textbackslash mathbf\{e\}\_i overlaid. (d) Argmax map with value-transformed vectors \textbackslash mathbf\{v\}\_j overlaid.}]{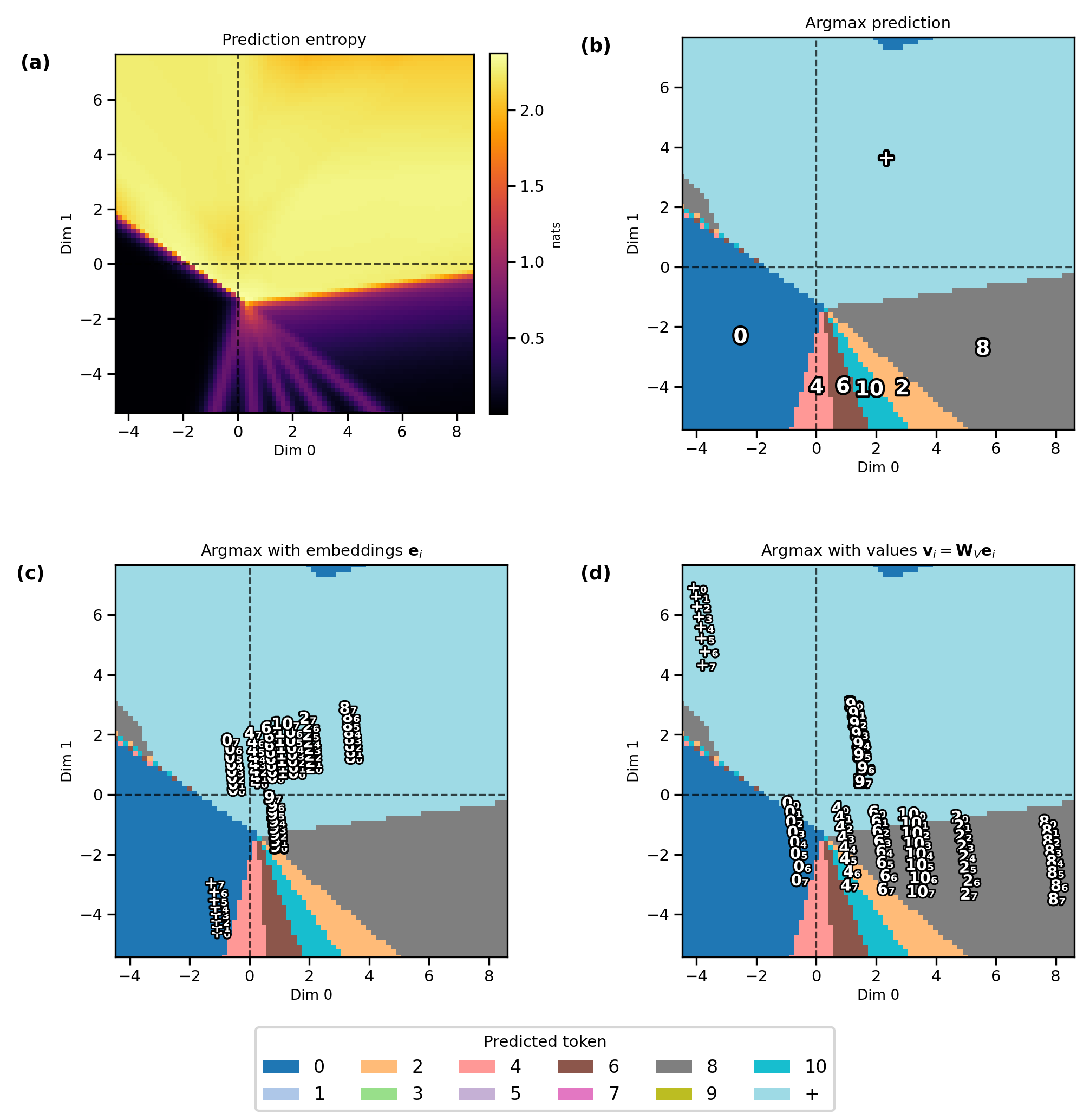}
\caption[\textbf{Figure 10.} Output landscape summary (summary of Figure
9). (a) Entropy of the output distribution (nats); high in the
indifference region (top half), near zero in even-number target zones
(bottom half). (b) Argmax prediction map: color and annotation indicate
which token has the highest predicted probability at each point. (c)
Argmax map with combined embeddings \(\mathbf{e}_i\) overlaid. (d)
Argmax map with value-transformed vectors \(\mathbf{v}_j\)
overlaid.]{\textbf{Figure 10.} Output landscape summary (summary of
Figure 9). (a) Entropy of the output distribution (nats); high in the
indifference region (top half), near zero in even-number target zones
(bottom half). (b) Argmax prediction map: color and annotation indicate
which token has the highest predicted probability at each point. (c)
Argmax map with combined embeddings \(\mathbf{e}_i\) overlaid. (d)
Argmax map with value-transformed vectors \(\mathbf{v}_j\) overlaid.}
\end{figure}

\subsection{Tracing a Sequence Through the Pipeline}\label{tracing-a-sequence-through-the-pipeline}

The preceding analysis characterized the model's learned parameters in
the abstract --- all 96 possible token+position combinations. We now
ground this analysis by tracing a single concrete sequence,
\texttt{4\ 1\ +\ 4\ 6\ 9\ 5\ +}, through the complete inference pipeline
and verifying that each step works as expected. This sequence ends in a
\texttt{+} to illustrate next token prediction. The model's output for
the next token in the sequence should be a 6.

\textbf{Embedding.} The embeddings for this sequence are a subset of the
total set of token+position embeddings, as there can only be one token
in each position. Thus, this sequence uses only 8 out of the 96 total
possible token+position combinations. In Figure 11, we show the tokens
for this sequence against a backdrop of all 96 possible token+position
combinations as shown in Figure 5. As we showed earlier, the even, odd
and \texttt{+} token embeddings are separated from each other in the
embedding space, and the position of each token in the sequence
determines where it appears in the ``ladder'' structure for that token.

\begin{figure}
\centering
\pandocbounded{\includegraphics[keepaspectratio,alt={Figure 11. Embeddings for the demo sequence 4 1 + 4 6 9 5 +. a--c: Embedding heatmaps for tokens, positions and token+position. d--f: 2D scatter plot views of tokens, positions, and token+position combinations, shown over a backdrop of all possible tokens, positions, and token+position embeddings (light gray). Insets in (d) and (f) magnify clusters of odd digits.}]{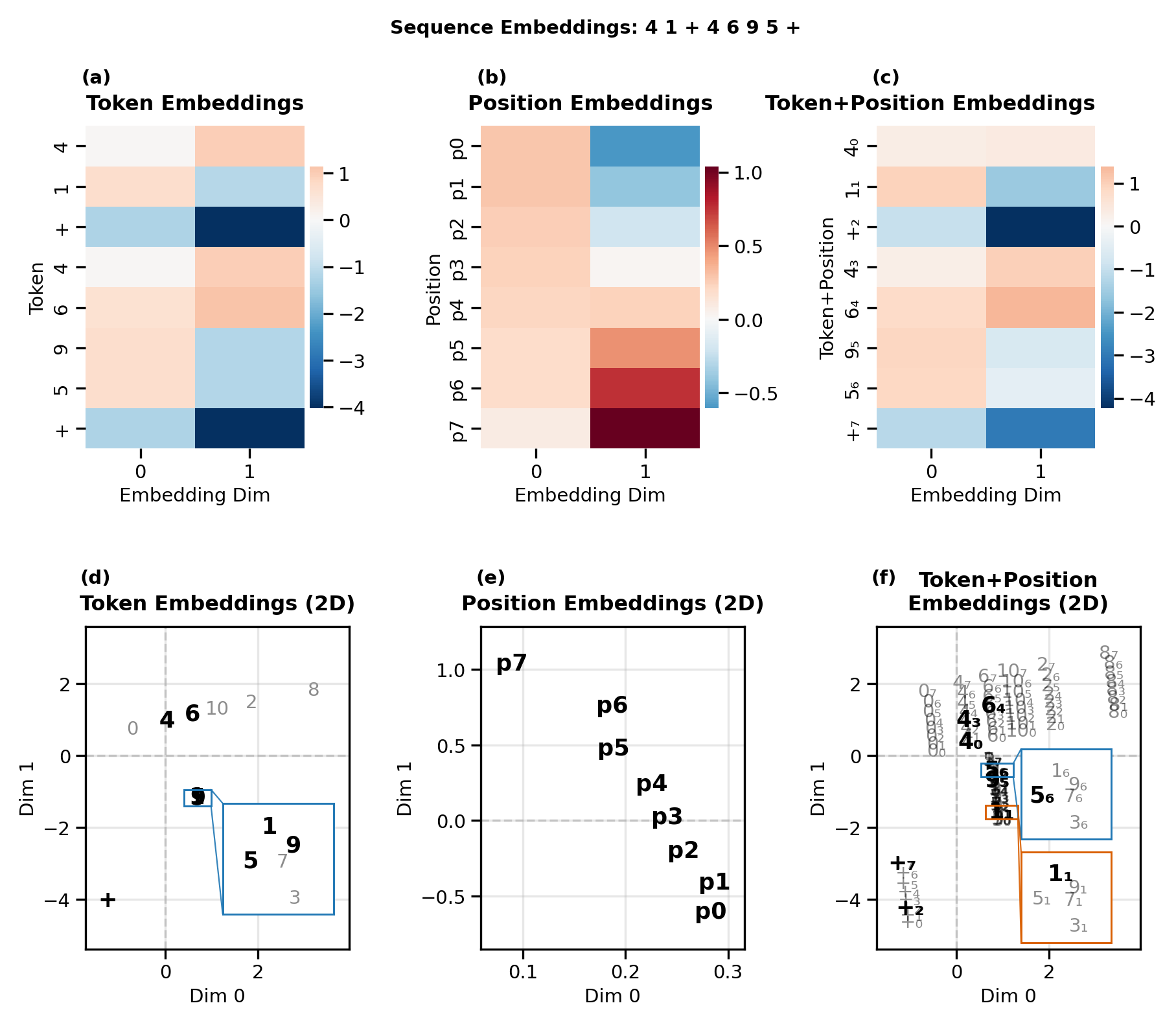}}
\caption[\textbf{Figure 11.} Embeddings for the demo sequence
\protect\texttt{4\ 1\ +\ 4\ 6\ 9\ 5\ +}. \textbf{a--c:} Embedding
heatmaps for tokens, positions and token+position. \textbf{d--f:} 2D
scatter plot views of tokens, positions, and token+position
combinations, shown over a backdrop of all possible tokens, positions,
and token+position embeddings (light gray). Insets in (d) and (f)
magnify clusters of odd digits.]{\textbf{Figure 11.} Embeddings for the
demo sequence \protect\texttt{4\ 1\ +\ 4\ 6\ 9\ 5\ +}. \textbf{a--c:}
Embedding heatmaps for tokens, positions and token+position.
\textbf{d--f:} 2D scatter plot views of tokens, positions, and
token+position combinations, shown over a backdrop of all possible
tokens, positions, and token+position embeddings (light gray). Insets in
(d) and (f) magnify clusters of odd digits.}
\end{figure}

\textbf{Attention.} The model must compute the attention matrix for this
sequence by forming
\(\mathbf{Q}, \mathbf{K} \in \mathbb{R}^{T \times d_k}\) for the \(T\)
positions and taking dot products between query rows and key rows
(i.e.~the matrix \(\mathbf{Q}\mathbf{K}^\top\)). Figure 12 shows the Q
and K heatmaps for the representation of the tokens within this
sequence, and panel (c) shows these Q and K embeddings on the same
scatterplot against a backdrop of all 96 possible queries and keys, as
in Figure 7.

\begin{figure}
\centering
\pandocbounded{\includegraphics[keepaspectratio,alt={Figure 12. Q and K for the demo sequence. (a) Q heatmap, (b) K heatmap, (c) scatterplot of queries (blue) and keys (red) for tokens within our test sequence against a backdrop of all 96 possible queries and keys (gray).}]{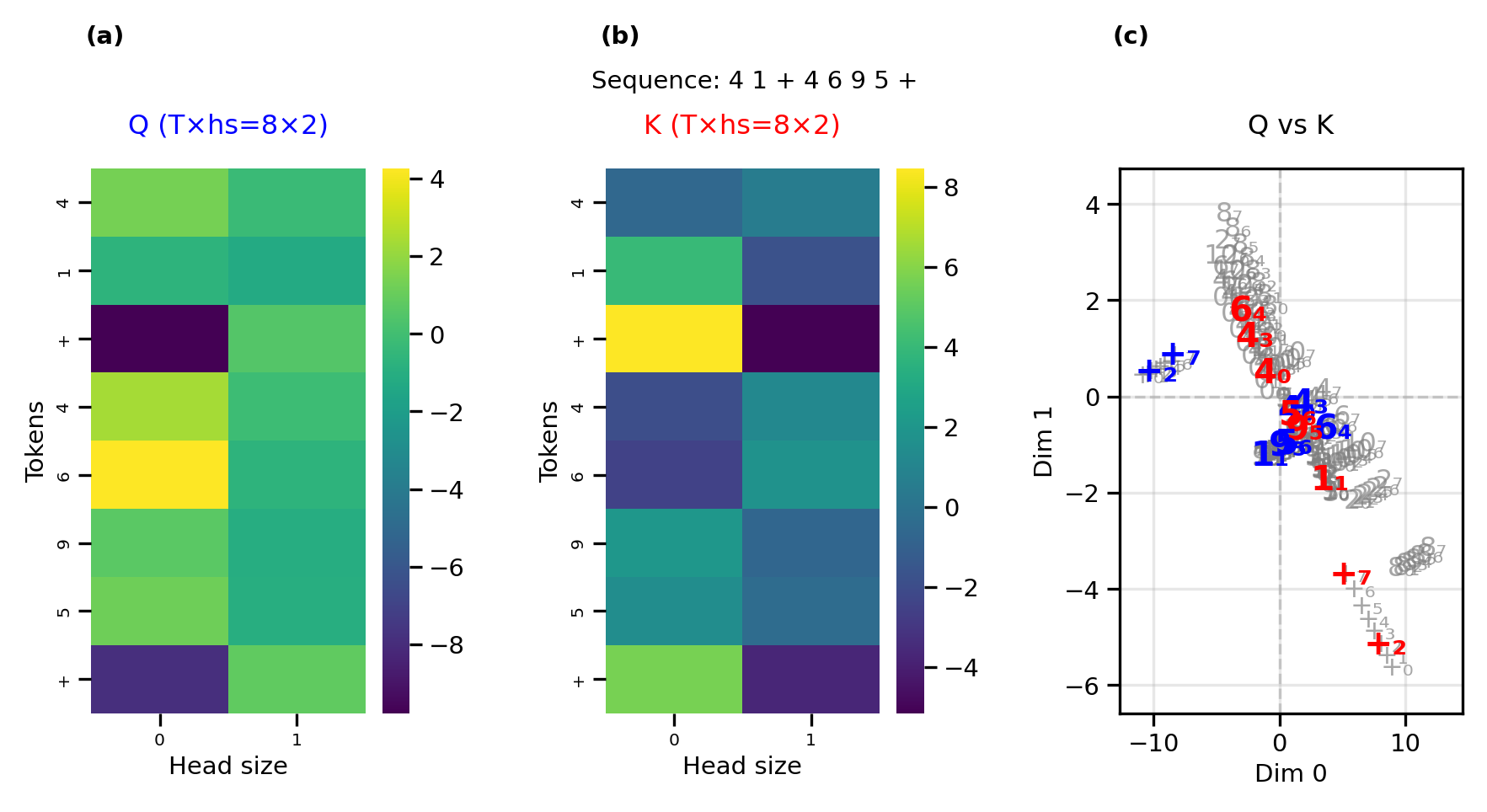}}
\caption[\textbf{Figure 12.} Q and K for the demo sequence. (a) Q
heatmap, (b) K heatmap, (c) scatterplot of queries (blue) and keys (red)
for tokens within our test sequence against a backdrop of all 96
possible queries and keys (gray).]{\textbf{Figure 12.} Q and K for the
demo sequence. (a) Q heatmap, (b) K heatmap, (c) scatterplot of queries
(blue) and keys (red) for tokens within our test sequence against a
backdrop of all 96 possible queries and keys (gray).}
\end{figure}

For each token within the sequence, the model should compute the dot
product between that token's query and the key of every token in the
sequence. To illustrate this, for each query, we show what the dot
product with that query would be at an arbitrary point in space. We then
overlay the actual keys within our test sequence over this space for
each query, making it clear which keys would produce the largest dot
product. For each query, we also gray out the keys that come in later
positions to show the effect of the causal masking (Figure 13a).

It is apparent from these visualizations that once causal masking is
applied, the dot product between the query of each \texttt{+} token and
the key of the most recent even number relative to that \texttt{+} token
will be the largest for that query relative to other unmasked keys. We
can verify this by looking at the heatmap of Figure 13b, which shows the
masked query-key dot products. If we look at the rows for the two
\texttt{+} tokens that appear in the sequence, each of those rows has
the largest value in the column belonging to the most recent even number
relative to that \texttt{+} token. Once we generate the final attention
matrix via normalization by \({\sqrt{d_k}}\) and applying softmax, it is
clear that the \texttt{+} tokens attend to their respective most recent
even numbers (Figure 13c).

\begin{figure}
\centering
\pandocbounded{\includegraphics[keepaspectratio,alt={Figure 13. Attention computation for the demo sequence. (a) The background of each panel displays the dot product of a specific query (blue) in the test sequence with every point in space (green = high, white = low). The unmasked (red) and masked (gray) keys are overlaid on this space to show their dot products with the selected query. (b) The masked \textbackslash mathbf\{Q\}\textbackslash mathbf\{K\}\^{}\textbackslash top matrix, (c) the attention matrix after normalization by \{\textbackslash sqrt\{d\_k\}\} and applying softmax.}]{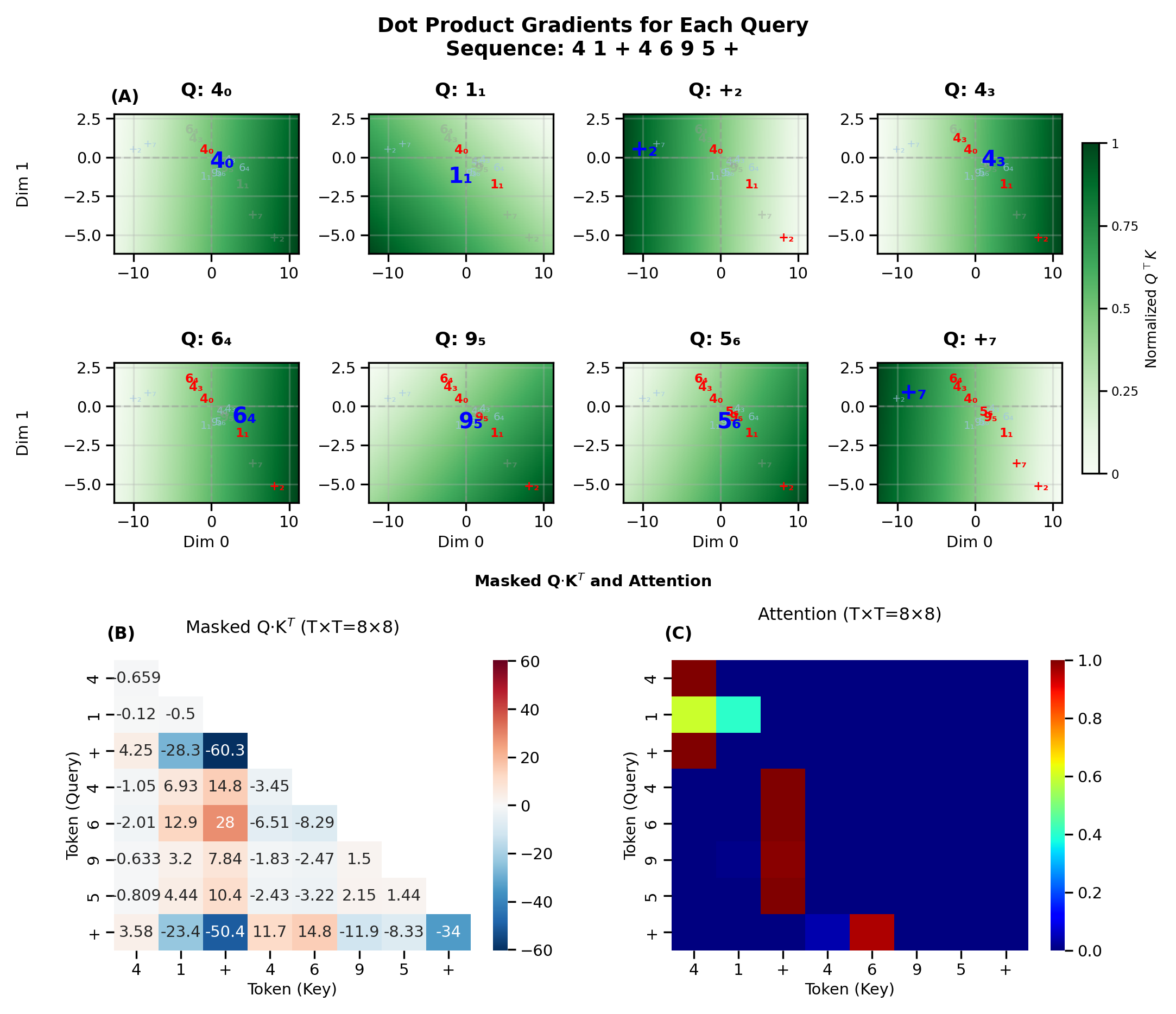}}
\caption[\textbf{Figure 13.} Attention computation for the demo
sequence. (a) The background of each panel displays the dot product of a
specific query (blue) in the test sequence with every point in space
(green = high, white = low). The unmasked (red) and masked (gray) keys
are overlaid on this space to show their dot products with the selected
query. (b) The masked \(\mathbf{Q}\mathbf{K}^\top\) matrix, (c) the
attention matrix after normalization by \({\sqrt{d_k}}\) and applying
softmax.]{\textbf{Figure 13.} Attention computation for the demo
sequence. (a) The background of each panel displays the dot product of a
specific query (blue) in the test sequence with every point in space
(green = high, white = low). The unmasked (red) and masked (gray) keys
are overlaid on this space to show their dot products with the selected
query. (b) The masked \(\mathbf{Q}\mathbf{K}^\top\) matrix, (c) the
attention matrix after normalization by \({\sqrt{d_k}}\) and applying
softmax.}
\end{figure}

\textbf{Value routing.} Now that we have the attention weight matrix
\(\mathbf{A}\), we need to apply it to the value vectors in order to
produce the attention vector for each token
\(\mathrm{Attn}(\mathbf{E})_i\). Specifically, the attention vectors are
the attention-weighted sum of the value vectors for each token,
i.e.~\(\mathrm{Attn}(\mathbf{E})_i = \sum_{j=0}^{i} \alpha_{ij} \mathbf{v}_j\).
We can generate a matrix whose rows are the attention outputs for each
position by multiplying the attention weight matrix \(\mathbf{A}\)
(Figure 14a) by the value matrix \(\mathbf{V}\) (Figure 14b) such that
we have
\(\mathrm{Attn}(\mathbf{E})_i^\top = (\mathbf{A} \mathbf{V})_{i,:}\)
(Figure 14c). We also show the value vectors \(\mathbf{v}_j\) (Figure
14d) and attention vectors \(\mathrm{Attn}(\mathbf{E})_i\) (Figure 14e)
as scatter plots. The attention weights for each token here effectively
select value vectors to be used for the prediction; the attention
vectors for the \texttt{+} tokens, for example, are basically copies of
the value vectors of the most recent even number.

\begin{figure}
\centering
\pandocbounded{\includegraphics[keepaspectratio,alt={Figure 14. Values and attention output for the demo sequence. (a) Attention weight matrix as in Figure 13c, (b) Values for each digit shown as heatmaps. (c) Attention outputs shown as heatmaps. (d) Values shown as scatter plot. (e) Attention output shown as scatter plot. Inset: zoom-in.}]{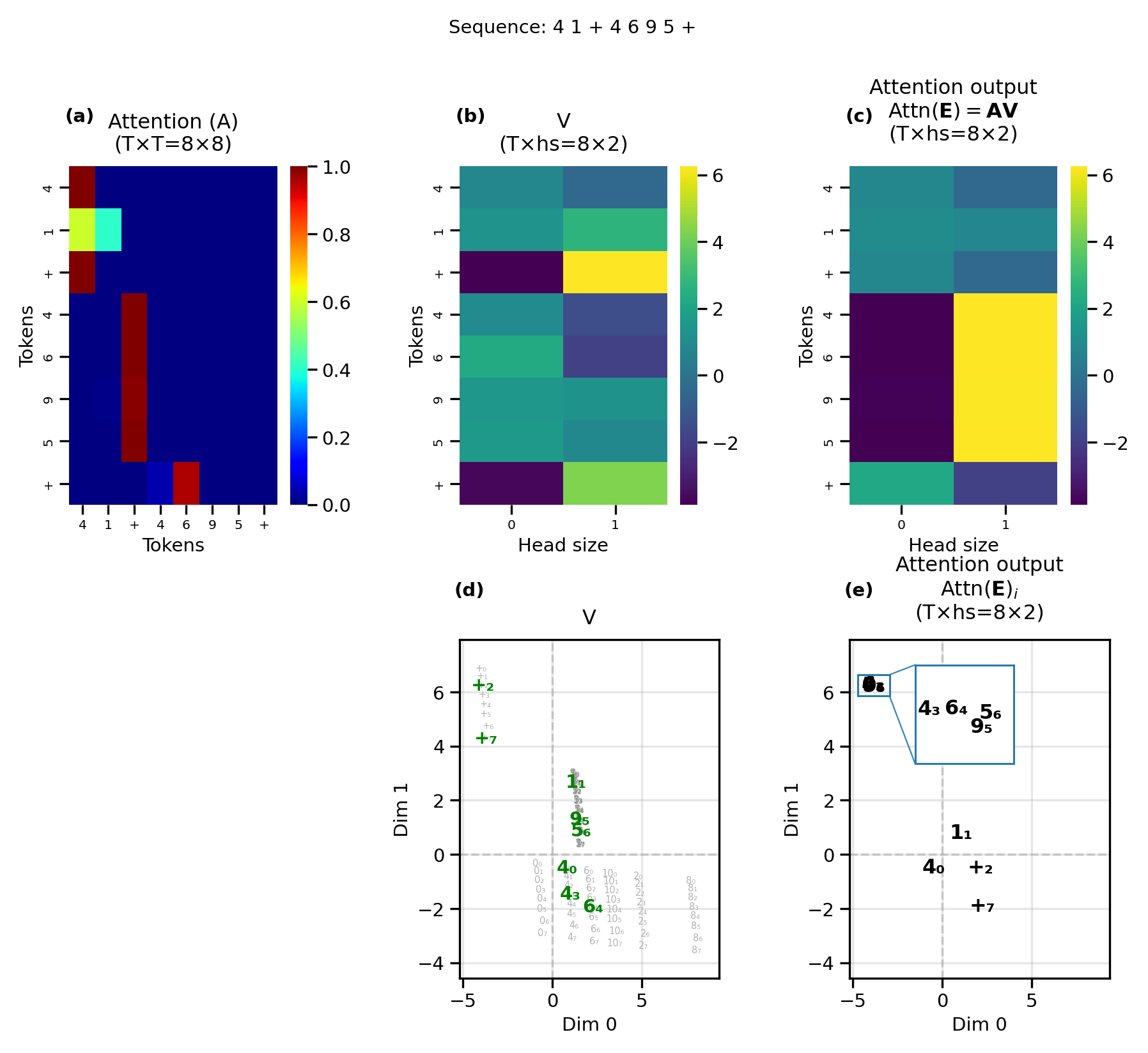}}
\caption[\textbf{Figure 14.} Values and attention output for the demo
sequence. (a) Attention weight matrix as in Figure 13c, (b) Values for
each digit shown as heatmaps. (c) Attention outputs shown as heatmaps.
(d) Values shown as scatter plot. (e) Attention output shown as scatter
plot. Inset: zoom-in.]{\textbf{Figure 14.} Values and attention output
for the demo sequence. (a) Attention weight matrix as in Figure 13c, (b)
Values for each digit shown as heatmaps. (c) Attention outputs shown as
heatmaps. (d) Values shown as scatter plot. (e) Attention output shown
as scatter plot. Inset: zoom-in.}
\end{figure}

\textbf{Predicting the next token.} With the attention vectors
\(\mathrm{Attn}(\mathbf{E})_i\) for each embedding in hand, we are ready
to predict the next token for each token in our test sequence. For the
token at position \(i\) in the sequence, we first compute
\(\mathbf{z}_i = \mathbf{e}_i + \mathrm{Attn}(\mathbf{E})_i\). In other
words, we add the token's attention vector to its embedding and see
where it lands in the output space. We can think of this addition as the
original token+position embedding being nudged to the right zone in the
output space by the attention vector. By showing the input embeddings,
attention vectors, and their sums as both heatmaps and scatter plots on
the background of the output prediction space (Figure 15), we observe
that the embeddings for all digits are nudged toward the high-entropy
region where all tokens are somewhat likely to occur, but the \texttt{+}
token is the most likely prediction. The \texttt{+} embeddings, however,
are nudged to the regions of the correct predictions for the most recent
even number. The \texttt{+} in position 2 correctly predicts that the
next number in the sequence should be a 4. The final token in the
sequence, the \texttt{+} in position 7, correctly predicts that the next
number in the sequence should be a 6, thus demonstrating that the
transformer can correctly produce new tokens that are beyond the end of
the input sequence.

\begin{figure}
\centering
\pandocbounded{\includegraphics[keepaspectratio,alt={Figure 15. Predicting the next token by summing the input embeddings and attention vectors. (a) Top: Heatmaps illustrating the 2D values for the input embeddings (\textbackslash mathbf\{E\}) for each token in the test sequence. Bottom: Input embeddings in the 2D output space, where the background colors represent the most likely next-token prediction. (b) Top: The attention vectors (\textbackslash mathrm\{Attn\}(\textbackslash mathbf\{E\})) as heatmap. Bottom: The attention vectors shown as arrows from the origin. (c) Top: The sum of the input embeddings and the attention vectors (\textbackslash mathbf\{Z\}) as heatmap. Bottom: Input embeddings are moved by the attention vectors (arrows with annotated heads) to the correct region for predicting the next token.}]{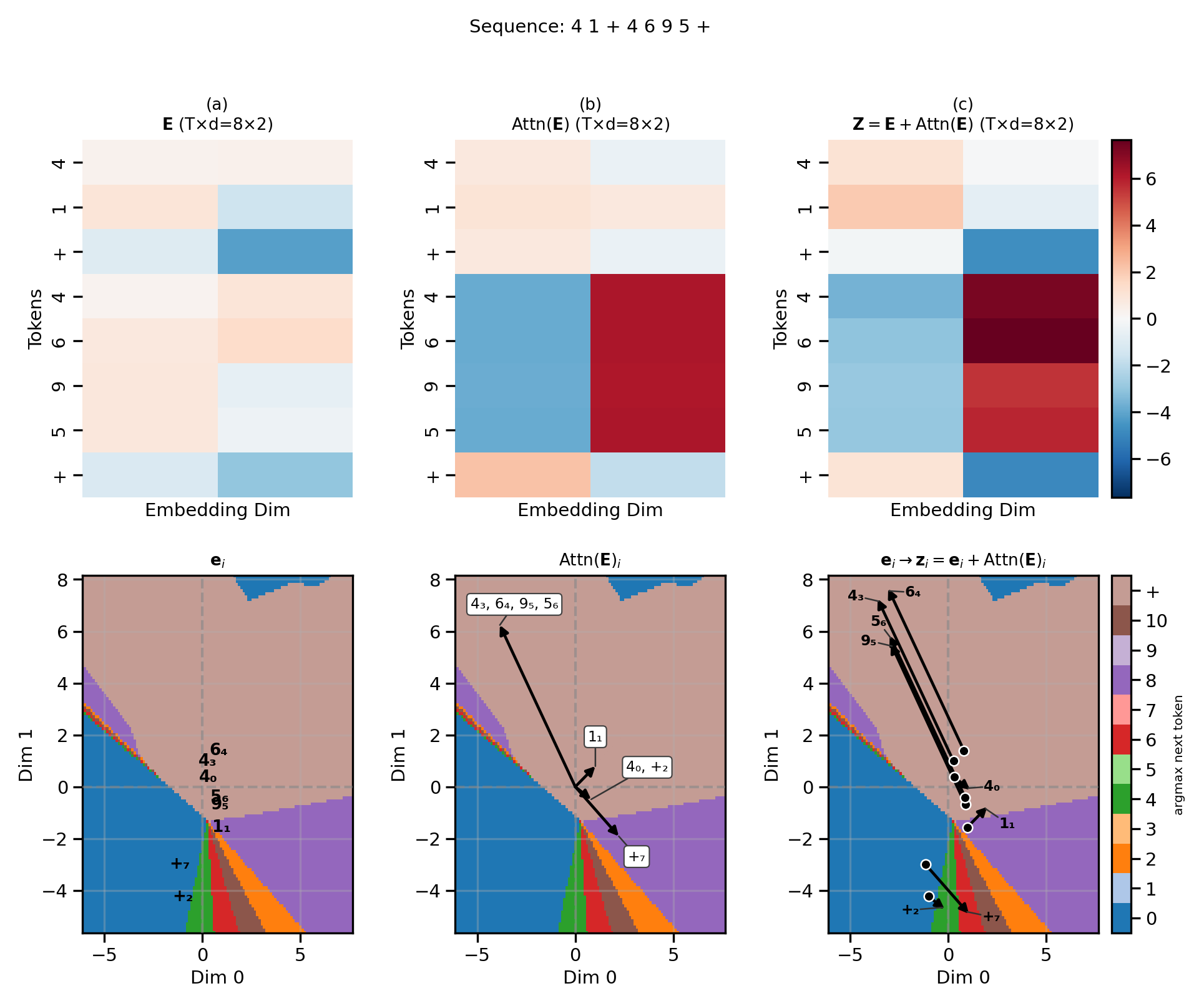}}
\caption[\textbf{Figure 15. Predicting the next token by summing the
input embeddings and attention vectors.} (a) Top: Heatmaps illustrating
the 2D values for the input embeddings (\(\mathbf{E}\)) for each token
in the test sequence. Bottom: Input embeddings in the 2D output space,
where the background colors represent the most likely next-token
prediction. (b) Top: The attention vectors
(\(\mathrm{Attn}(\mathbf{E})\)) as heatmap. Bottom: The attention
vectors shown as arrows from the origin. (c) Top: The sum of the input
embeddings and the attention vectors (\(\mathbf{Z}\)) as heatmap.
Bottom: Input embeddings are moved by the attention vectors (arrows with
annotated heads) to the correct region for predicting the next
token.]{\textbf{Figure 15. Predicting the next token by summing the
input embeddings and attention vectors.} (a) Top: Heatmaps illustrating
the 2D values for the input embeddings (\(\mathbf{E}\)) for each token
in the test sequence. Bottom: Input embeddings in the 2D output space,
where the background colors represent the most likely next-token
prediction. (b) Top: The attention vectors
(\(\mathrm{Attn}(\mathbf{E})\)) as heatmap. Bottom: The attention
vectors shown as arrows from the origin. (c) Top: The sum of the input
embeddings and the attention vectors (\(\mathbf{Z}\)) as heatmap.
Bottom: Input embeddings are moved by the attention vectors (arrows with
annotated heads) to the correct region for predicting the next token.}
\end{figure}

\section{Geometry as Algorithm: Summary}\label{geometry-as-algorithm-summary}

For the plus-last-even rule, the model's behavior decomposes into a
six-step algorithm. Each step corresponds to a specific geometric
structure visible in the preceding figures.

\begin{enumerate}
\def\labelenumi{\arabic{enumi}.}
\item
  \textbf{Encode.} Map each (token, position) pair to a 2D point: token
  embedding \(\mathbf{x}_i\) plus position embedding \(\mathbf{p}_i\)
  gives the combined embedding
  \(\mathbf{e}_i = \mathbf{x}_i + \mathbf{p}_i\). The embedding scatter
  plots (Figure 5) show this mapping, with even numbers, odd numbers,
  and \texttt{+} occupying distinct regions.
\item
  \textbf{Detect the \texttt{+}.} The query projection \(\mathbf{W}_Q\)
  maps the \texttt{+} embedding to a query vector that is geometrically
  distinct from number queries (Figure 7, top).
\item
  \textbf{Retrieve the last even number.} The dot product between the
  \texttt{+} query and all keys (produced by the \(\mathbf{W}_K\)
  projection) yields high attention weight for even-digit keys, with
  recency encoded in the positional component of the key layout. The
  focused-query analysis (Figure 7, bottom) and full dot product matrix
  for \texttt{+} queries (Figure 8) make this retrieval pattern
  explicit; the per-query gradient figure (Figure 13) shows it for the
  demo sequence.
\item
  \textbf{Produce attention vectors.} Multiplying the attention matrix
  \(\mathbf{A}\) by the value matrix \(\mathbf{V}\) (generated by the
  \(\mathbf{W}_V\) projection) produces a weighted sum of value vectors
  for each token. For the \texttt{+} tokens, attention selects the value
  vector of the most recent even-digit position.
\item
  \textbf{Use attention vectors to push embeddings into the correct
  output zone.} For each token, the attention vectors are added to the
  original token+position embedding from the residual in order to nudge
  that token's representation into the correct zone for the prediction
  of the next token. In the case of the \texttt{+} tokens, the
  \texttt{+} embeddings are nudged in the direction of the zone of the
  most recent even number.
\item
  \textbf{Predict the next even number.} The output network maps the
  shifted representation \(\mathbf{z}_i\) to a probability distribution
  over the vocabulary. Because step 5 places each \texttt{+}
  representation inside the target zone of the most recent even number,
  that even number receives the highest probability and is predicted as
  the next token (Figures 9, 10, and 15).
\end{enumerate}

We have thus demonstrated how the representational geometry of the
transformer (Vaswani et al., 2017) can be directly interpreted as an
algorithm that solves the plus-last-even task.

\section{Discussion}\label{discussion}

\textbf{Conclusions and implications.} We have shown that it is possible
to train a minimal transformer with an embedding size and head size of 2
to near-perfect accuracy by selecting a sufficiently simple goal, namely
the plus-last-even task. The model's low dimensionality enables us to
inspect the learned internal information geometry of the transformer at
every step in the architecture. This approach enables near-complete
interpretability of the attention and embedding pathway for this minimal
model and task. Rather than only observing what the model predicts, we
can trace exactly how the computation is organized inside the network
and understand how the algorithm is implemented without the need for
dimensionality reduction.

\textbf{Limitations.} The primary limitation of this approach is the
representational capacity of the 2D space. While \(\mathbb{R}^2\) is
sufficient for the 12-token plus-last-even task, it would not be
effective for more complex conditional rules, and certainly not for
full-fledged AI tasks such as language modeling. Furthermore, this study
is restricted to a single-layer, single-head architecture. In deeper
models with more attention heads, the interaction between successive
residual updates and FFN transformations introduces topological
complexities that are harder to interpret as a single-step geometric
nudge toward a decision boundary. In addition, our analysis is based on
a single trained model and default random seed; a more robust analysis
would look at variation in geometric representations across seeds.

\textbf{Future directions.} The plus-last-even task is an artificial
task that was chosen for pedagogical purposes to illustrate various
facets of the transformer computation. Additional simple tasks,
including basic mathematical or logical operations, can potentially also
be solved using our minimal transformer architecture. The internal
representations learned when solving different tasks can provide further
insight into how transformers use information geometry to approach
different kinds of problems. Moreover, transformers trained on the same
tasks but with different random seeds can provide insight into what
aspects of the information geometry are necessary to solve a task versus
what aspects are left up to the model's ``creative discretion''.

Although our work emphasized using 2-dimensional latent spaces for the
purpose of direct, lossless visualization, the visualization approaches
that we used here can potentially be applied to more complex models if
dimensionality reduction techniques are used and steps are taken to
isolate the representations at each layer. However, important
information is often lost in the dimensionality reduction process,
limiting the utility of such an approach.

Neuroscientists have become increasingly interested in relating the
representational geometry of complex perceptions and behaviors in the
brain to those found in large language models (e.g., Caucheteux et al.,
2021; Hosseini et al., 2024; Sun et al., 2025; Doerig et al., 2025). Our
framework allows for the possibility of exploring representational
geometry in sufficiently simple tasks that can also be used in
experimental neuroscience, potentially enabling a direct comparison of
neural activity to a transformer's internal representations.

\section{Movies}\label{movies}

The following animations show the evolution of the model's learned
geometry over the course of training (one frame per checkpoint, 200
frames total). These are available as GIF/MP4 files in
\texttt{\seqsplit{plus\_last\_even/plots/learning\_dynamics/}} of the
code repository
(\url{https://github.com/Raneem-mahajne/creating_transformer/tree/main/plus_last_even/plots/learning_dynamics}).

\noindent

\begin{xltabular}{\linewidth}{@{}l>{\raggedright\arraybackslash}p{3.5cm}>{\raggedright\arraybackslash}X@{}}
\toprule
Movie & \texttt{File} & Description \\
\midrule
Movie 1 & \parbox[t]{3.5cm}{\raggedright\ttfamily\href{https://github.com/Raneem-mahajne/creating_transformer/blob/main/plus_last_even/plots/learning_dynamics/01_embeddings_scatterplots.gif}{\seqsplit{01\_embeddings\_scatterplots.gif}}} & Evolution of token, position, and combined embedding scatter plots. The \texttt{+} token separates from numbers first; the even/odd split solidifies by step 5,000--10,000. \\
Movie 2 & \parbox[t]{3.5cm}{\raggedright\ttfamily\href{https://github.com/Raneem-mahajne/creating_transformer/blob/main/plus_last_even/plots/learning_dynamics/05_output_heatmaps_with_embeddings.gif}{\seqsplit{05\_output\_heatmaps\_with\_embeddings.gif}}} & Co-evolution of the LM head's output probability landscape and embedding positions. Decision boundaries sharpen progressively from a uniform initialization. \\
Movie 3 & \parbox[t]{3.5cm}{\raggedright\ttfamily\href{https://github.com/Raneem-mahajne/creating_transformer/blob/main/plus_last_even/plots/learning_dynamics/02_embedding_qkv_comprehensive.gif}{\seqsplit{02\_embedding\_qkv\_comprehensive.gif}}} & Specialization of the Q, K, and V subspaces. All three projections are initially identical and develop distinct geometry as training progresses. \\
Movie 4 & \parbox[t]{3.5cm}{\raggedright\ttfamily\href{https://github.com/Raneem-mahajne/creating_transformer/blob/main/plus_last_even/plots/learning_dynamics/03_qk_embedding_space.gif}{\seqsplit{03\_qk\_embedding\_space.gif}}} & Separation of query and key subspaces. \texttt{+} queries migrate away from number queries; even-number keys align with the \texttt{+} query direction. \\
Movie 5 & \parbox[t]{3.5cm}{\raggedright\ttfamily\href{https://github.com/Raneem-mahajne/creating_transformer/blob/main/plus_last_even/plots/learning_dynamics/04_qk_space_plus_attention.gif}{\seqsplit{04\_qk\_space\_plus\_attention.gif}}} & Evolution of the full attention matrix alongside the Q/K scatter. The \texttt{+}-row entries concentrate on even-number columns over training. \\
\bottomrule
\end{xltabular}

\subsection{Supplementary Figures}\label{supplementary-figures}

Additional static figures in
\texttt{\seqsplit{plus\_last\_even/plots/supplementary/}},
\texttt{\seqsplit{plus\_last\_even/plots/extended/}}, and
\texttt{plus\_last\_even/plots/a4/}:

\noindent

\begin{xltabular}{\linewidth}{@{}l>{\raggedright\arraybackslash}p{3.2cm}>{\raggedright\arraybackslash}X@{}}
\toprule
Figure & \texttt{File} & Description \\
\midrule
S1 & \parbox[t]{3.2cm}{\raggedright\ttfamily\seqsplit{supplementary/07\_qkv\_overview.png}} & Comprehensive 3$\times$3 panel: token, position, and combined embeddings; Q/K/V transformed spaces; Q+K overlay; and attention output. \\
S2 & \parbox[t]{3.2cm}{\raggedright\ttfamily\seqsplit{supplementary/14\_attention\_matrix.png}} & Per-sequence attention matrices alongside LM head linear input, logits, and output probabilities for three demo sequences. \\
S3 & \parbox[t]{3.2cm}{\raggedright\ttfamily\seqsplit{supplementary/16\_value\_arrows.png}} & Value vectors (original, transformed, residual) for three demo sequences, with correctness indicated by green/red markers. \\
S4 & \parbox[t]{3.2cm}{\raggedright\ttfamily\seqsplit{extended/08\_qkv\_transforms\_extended.png}} & Extended QKV figure with per-dimension heatmaps (tokens$\times$positions) for Q, K, and V. \\
S5 & \parbox[t]{3.2cm}{\raggedright\ttfamily\seqsplit{a4/11\_qk\_full\_heatmap.png}} & Full 96×96 pre-softmax query–key score matrix \(\mathbf{Q}\mathbf{K}^\top\) (dot products), covering all query tokens and positions vs. all key tokens and positions (no masking/softmax), organized as a 12×12 grid of token–token blocks. \\
S6 & \parbox[t]{3.2cm}{\raggedright\ttfamily\seqsplit{a4/07\_output\_probs\_embed.png}} & Per-unit output probability heatmaps with all 96 combined embeddings $\mathbf{e}_i$ overlaid (token + position labels). \\
S7 & \parbox[t]{3.2cm}{\raggedright\ttfamily\seqsplit{a4/12\_probability\_heatmap\_with\_values.png}} & Per-unit output probability heatmaps with all 96 value-transformed vectors $\mathbf{W}_V \mathbf{e}_i$ overlaid. \\
\bottomrule
\end{xltabular}

\section{Reproducing the Results}\label{reproducing-the-results}

\textbf{Train and visualize:}

\begin{Shaded}
\begin{Highlighting}[]
\ExtensionTok{python}\NormalTok{ main.py plus\_last\_even}
\end{Highlighting}
\end{Shaded}

\textbf{Visualize from an existing checkpoint:}

\begin{Shaded}
\begin{Highlighting}[]
\ExtensionTok{python}\NormalTok{ main.py plus\_last\_even }\AttributeTok{{-}{-}visualize}
\end{Highlighting}
\end{Shaded}

\textbf{Visualize a specific training step:}

\begin{Shaded}
\begin{Highlighting}[]
\ExtensionTok{python}\NormalTok{ main.py plus\_last\_even }\AttributeTok{{-}{-}visualize} \AttributeTok{{-}{-}step}\NormalTok{ 5000}
\end{Highlighting}
\end{Shaded}

\textbf{Generate learning-dynamics videos:}

\begin{Shaded}
\begin{Highlighting}[]
\ExtensionTok{python}\NormalTok{ main.py plus\_last\_even }\AttributeTok{{-}{-}video}
\ExtensionTok{python}\NormalTok{ main.py plus\_last\_even }\AttributeTok{{-}{-}video{-}qkv}
\end{Highlighting}
\end{Shaded}

\textbf{Dependencies:} PyTorch, NumPy, Matplotlib, Pillow, imageio.

\section*{Declaration on the use of artificial intelligence}\label{declaration-on-the-use-of-artificial-intelligence}
\addcontentsline{toc}{section}{Declaration on the use of artificial intelligence}

Nearly all code --- including the transformer model, training pipeline,
and figure/video generation --- was produced with Cursor Agent. All
training outcomes, figures, and interpretive claims were independently
verified by the authors. The first draft of this paper was written by
Claude Opus 4.6; the entirety of the draft was thoroughly rewritten,
edited, and revised by the authors for factual accuracy, clarity, and
language. Some later edits to the figures, the figure-generation code,
and the text were made with Claude Opus 5.5.

\section*{Acknowledgments}\label{acknowledgments}
\addcontentsline{toc}{section}{Acknowledgments}

We thank Idan Segev for his guidance and support throughout this
project. This work received generous support from the Drahi Family
Foundation and the Gatsby Charitable Foundation.

\section*{References}\label{references}
\addcontentsline{toc}{section}{References}

\begingroup\raggedright\setlist[itemize]{label=\textbullet,leftmargin=1.6em,itemsep=0.4em,topsep=0.5em,parsep=0.12em}

\begin{itemize}
\item
  Belrose, N., Ostrovsky, I., McKinney, L., Furman, Z., Smith, L.,
  Halawi, D., Biderman, S., \& Steinhardt, J. (2023). Eliciting latent
  predictions from transformers with the tuned lens. \emph{arXiv
  preprint} arXiv:2303.08112. https://arxiv.org/abs/2303.08112
\item
  Bricken, T., Templeton, A., Batson, J., Chen, B., Jermyn, A., et
  al.~(2023). Towards monosemanticity: Decomposing language models with
  dictionary learning. \emph{Transformer Circuits Thread}.
  https://transformer-circuits.pub/2023/monosemantic-features
\item
  Caucheteux, C., Gramfort, A., \& King, J.-R. (2021). GPT-2's
  activations predict the degree of semantic comprehension in the human
  brain. \emph{bioRxiv}. https://doi.org/10.1101/2021.04.20.440622
\item
  Clark, K., Khandelwal, U., Levy, O., \& Manning, C. D. (2019). What
  does BERT look at? An analysis of BERT's attention. \emph{ACL Workshop
  on BlackboxNLP}. https://arxiv.org/abs/1906.04341
\item
  Dar, G., Geva, M., Gupta, A., \& Berant, J. (2023). Analyzing
  transformers in embedding space. \emph{ACL}, 16124--16170.
  https://aclanthology.org/2023.acl-long.893/
\item
  Doerig, A., Kietzmann, T. C., Allen, E., et al.~(2025). High-level
  visual representations in the human brain are aligned with large
  language models. \emph{Nature Machine Intelligence}, \emph{7},
  1220--1234. https://doi.org/10.1038/s42256-025-01072-0
\item
  Elhage, N., Nanda, N., Olsson, C., et al.~(2021). A mathematical
  framework for transformer circuits. \emph{Transformer Circuits
  Thread}. https://transformer-circuits.pub/2021/framework/index.html
\item
  Elhage, N., Hume, T., Olsson, C., Schiefer, N., et al.~(2022). Toy
  models of superposition. \emph{Transformer Circuits Thread}.
  https://transformer-circuits.pub/2022/toy\_model/index.html
\item
  Friedman, D., Wettig, A., \& Chen, D. (2023). Learning Transformer
  Programs. \emph{NeurIPS}. https://arxiv.org/abs/2306.01128
\item
  Geva, M., Schuster, R., Berant, J., \& Levy, O. (2021). Transformer
  feed-forward layers are key-value memories. \emph{EMNLP}, 5484--5495.
  https://aclanthology.org/2021.emnlp-main.446/
\item
  Geva, M., Caciularu, A., Wang, K., \& Goldberg, Y. (2022). Transformer
  feed-forward layers build predictions by promoting concepts in the
  vocabulary space. \emph{EMNLP}, 30--45.
  https://aclanthology.org/2022.emnlp-main.3/
\item
  Gromov, A. (2023). Grokking modular arithmetic. \emph{arXiv preprint}
  arXiv:2301.02679v1. https://arxiv.org/pdf/2301.02679
\item
  Hanna, M., Liu, O., \& Variengien, A. (2023). How does GPT-2 compute
  greater-than?: Interpreting mathematical abilities in a pre-trained
  language model. \emph{NeurIPS}. https://arxiv.org/abs/2305.00586
\item
  Hosseini, E. A., Schrimpf, M., Zhang, Y., Bowman, S., Zaslavsky, N.,
  \& Fedorenko, E. (2024). Artificial neural network language models
  predict human brain responses to language even after a developmentally
  realistic amount of training. \emph{Neurobiology of Language},
  \emph{5}(1), 43--63. https://doi.org/10.1162/nol\_a\_00137
\item
  Li, K., Hopkins, A. K., Bau, D., Viégas, F., Pfister, H., \&
  Wattenberg, M. (2023). Emergent world representations: Exploring a
  sequence model trained on a synthetic task. \emph{ICLR}.
  https://arxiv.org/abs/2210.13382
\item
  Lindner, D., Kramár, J., Farquhar, S., Rahtz, M., McGrath, T., \&
  Mikulik, V. (2023). Tracr: Compiled transformers as a laboratory for
  interpretability. \emph{NeurIPS}. https://arxiv.org/abs/2301.05062
\item
  Liu, Z., Kitouni, O., Nolte, N., Michaud, E. J., Tegmark, M., \&
  Williams, M. (2022). Towards understanding grokking: An effective
  theory of representation learning. \emph{arXiv preprint}
  arXiv:2205.10343v2. https://arxiv.org/pdf/2205.10343
\item
  McInnes, L., Healy, J., \& Melville, J. (2018). UMAP: Uniform Manifold
  Approximation and Projection for Dimension Reduction. \emph{arXiv
  preprint} arXiv:1802.03426. https://arxiv.org/abs/1802.03426
\item
  Musat, T. (2024). Clustering and alignment: Understanding the training
  dynamics in modular addition. \emph{arXiv preprint}
  arXiv:2408.09414v2. https://arxiv.org/abs/2408.09414
\item
  Nanda, N., Chan, L., Lieberum, T., Smith, J., \& Steinhardt, J.
  (2023a). Progress measures for grokking via mechanistic
  interpretability. \emph{arXiv preprint} arXiv:2301.05217v1.
  https://arxiv.org/pdf/2301.05217v1
\item
  Nanda, N., Lee, A., \& Wattenberg, M. (2023b). Emergent linear
  representations in world models of self-supervised sequence models.
  \emph{BlackboxNLP}. https://aclanthology.org/2023.blackboxnlp-1.2/
\item
  nostalgebraist. (2020). Interpreting GPT: The logit lens. LessWrong.
  https://www.lesswrong.com/posts/AcKRB8wDpdaN6v6ru/interpreting-gpt-the-logit-lens
\item
  Pan, X., Philip, A., Xie, Z., \& Schwartz, O. (2024). Dissecting
  query-key interaction in vision transformers. \emph{NeurIPS},
  54595--54631.
  https://proceedings.neurips.cc/paper\_files/paper/2024/hash/6216515a5e0b3257c49dcb1647e497d1-Abstract.html
\item
  Park, K., Choe, Y. J., \& Veitch, V. (2024). The linear representation
  hypothesis and the geometry of large language models. \emph{ICML}.
  https://arxiv.org/abs/2311.03658
\item
  Power, A., Burda, Y., Edwards, H., Babuschkin, I., \& Misra, V.
  (2022). Grokking: Generalization beyond overfitting on small
  algorithmic datasets. \emph{arXiv preprint} arXiv:2201.02177.
  https://arxiv.org/abs/2201.02177
\item
  Quirke, P., \& Barez, F. (2024). Understanding addition in
  transformers. \emph{ICLR}. https://arxiv.org/abs/2310.13121
\item
  Radford, A., Narasimhan, K., Salimans, T., \& Sutskever, I. (2018).
  Improving language understanding by generative pre-training. OpenAI.
  https://cdn.openai.com/research-covers/language-unsupervised/language\_understanding\_paper.pdf
\item
  Stolfo, A., Belinkov, Y., \& Sachan, M. (2023). A mechanistic
  interpretation of arithmetic reasoning in language models using causal
  mediation analysis. \emph{EMNLP}, 7035--7052.
  https://aclanthology.org/2023.emnlp-main.435/
\item
  Sun, W., Winnubst, J., Natrajan, M., et al.~(2025). Learning produces
  an orthogonalized state machine in the hippocampus. \emph{Nature},
  \emph{640}, 165--175. https://doi.org/10.1038/s41586-024-08548-w
\item
  van der Maaten, L. \& Hinton, G. (2008). Visualizing data using t-SNE.
  \emph{JMLR}, 9, 2579--2605.
  https://www.jmlr.org/papers/v9/vandermaaten08a.html
\item
  Vaswani, A., Shazeer, N., Parmar, N., Uszkoreit, J., Jones, L., Gomez,
  A. N., Kaiser, Ł., \& Polosukhin, I. (2017). Attention is all you
  need. \emph{Advances in Neural Information Processing Systems}, 30.
  https://arxiv.org/abs/1706.03762
\item
  Vig, J. (2019). A multiscale visualization of attention in the
  transformer model. \emph{ACL System Demonstrations}.
  https://aclanthology.org/P19-3007/
\item
  Wang, K., Variengien, A., Conmy, A., Shlegeris, B., \& Steinhardt, J.
  (2022). Interpretability in the wild: A circuit for indirect object
  identification in GPT-2 small. \emph{ICLR}.
  https://arxiv.org/abs/2211.00593
\item
  Weiss, G., Goldberg, Y., \& Yahav, E. (2021). Thinking like
  transformers. \emph{ICML}.
  https://proceedings.mlr.press/v139/weiss21a.html
\item
  Welch Labs. (2025). \emph{The most complex model we actually
  understand} {[}Video{]}. YouTube.
  https://www.youtube.com/watch?v=D8GOeCFFby4
\item
  Zhong, Z., Liu, Z., Tegmark, M., \& Andreas, J. (2023). The clock and
  the pizza: Two stories in mechanistic explanation of neural networks.
  \emph{NeurIPS}. https://arxiv.org/abs/2306.17844
\item
  Zhou, H., Bradley, A., Littwin, E., Razin, N., Saremi, O., Susskind,
  J., Bengio, S., \& Nakkiran, P. (2024). What algorithms can
  Transformers learn? A study in length generalization. \emph{ICLR}.
  https://arxiv.org/abs/2310.16028
\end{itemize}

\endgroup

\end{document}